\documentclass[runningheads]{llncs}
\usepackage{graphicx}
\usepackage{comment}
\usepackage{amsmath,amssymb} 
\usepackage[dvipsnames]{xcolor}
\usepackage{xspace}

\usepackage{mathtools}
\usepackage{algorithm}
\usepackage{algorithmic}
\usepackage{nicefrac}

\newcommand{\M}[1]{\mathtt{#1}}
\newcommand{\V}[1]{\M{#1}}

\newcommand{\arr}[2]{\begin{array}{#1} #2\end{array}}
\newcommand{\mat}[2]{\left[\!\!\arr{#1}{#2}\!\!\right]}

\newcommand{\AS}[1]{{\color{red}AS: #1}}

\def\gb{Gr{\"o}bner basis\xspace}

\def\gbs{Gr{\"o}bner bases\xspace}

\begin{document}
\pagestyle{headings}
\mainmatter
\def\ECCVSubNumber{4407}  

\title{Minimal Rolling Shutter Absolute Pose with Unknown Focal Length and Radial Distortion} 

\addtolength{\baselineskip}{-0.25pt}
\titlerunning{RS Absolute Pose with Unknown Focal Length and Radial Distortion}
%
\author{Zuzana Kukelova\inst{1}\and
Cenek Albl\inst{2} \and
Akihiro Sugimoto\inst{4} \and
Konrad Schindler\inst{2} \and
Tomas Pajdla\inst{3}}
\authorrunning{Z. Kukelova et al.}
%
\institute{FEE - Faculty of Electrical Engineering, Czech Technical University in Prague\\
\email{kukelova@fel.cvut.cz}
\and
Photogrammetry and Remote Sensing, ETH Zurich, Switzerland\\
\email{cenek.albl@geod.baug.ethz.ch,schindler@ethz.ch}
\and
CIIRC - Czech Institute of Informatics, Robotics and Cybernetics, Czech Technical University in Prague \\
\email{pajdla@cvut.cz}
\and
National Institute of Informatics, Tokyo, Japan \\
\email{sugimoto@nii.ac.jp}
}

\maketitle

\begin{abstract}
The internal geometry of most modern consumer cameras is not adequately described by the perspective projection. Almost all cameras exhibit some radial lens distortion and are equipped with electronic rolling shutter that induces distortions when the camera moves during the image capture. When focal length has not been calibrated offline, the parameters that describe the radial and rolling shutter distortions are usually unknown. While for global shutter cameras, minimal solvers for the absolute camera pose and unknown focal length and radial distortion are available, solvers for the rolling shutter were missing. We present the first minimal solutions for the absolute pose of a rolling shutter camera with unknown rolling shutter parameters, focal length, and radial distortion. Our new minimal solvers combine iterative schemes  designed for calibrated rolling shutter cameras with fast generalized eigenvalue and \gb solvers. In a series of experiments, with both synthetic and real data, we show that our new solvers provide accurate estimates of the camera pose, rolling shutter parameters, focal length, and radial distortion parameters.

\end{abstract}
\section{Introduction}
Estimating the six degree-of-freedom (6DOF) pose of a camera is one of the fundamental problems in computer vision with many applications, including camera calibration~\cite{bouguet2008camera}, Structure-from-Motion (SfM)~\cite{snavely2006photo,schoenberger2016CVPR}, augmented reality (AR)~\cite{MicrosoftSpatialAnchors}, and visual localization~\cite{Sattler2017PAMI}.
The task is to compute the camera pose in the world coordinate system from 3D points in the world and their 2D projections in an image.

Solvers for the camera pose are usually used inside RANSAC-style hypothesis-and-test frameworks~\cite{Fischler1981}. For efficiency it is therefore important to employ \emph{minimal} solvers that generate the solution with a minimal number of point correspondences.
The minimal number of 2D-to-3D correspondences necessary to solve the absolute pose problem is three for a calibrated perspective camera. The earliest solver dates back  to 1841~\cite{Grunert-1841}. Since then, the problem has been revisited several times~\cite{Haralick1991,Ameller02camerapose,Fischler1981,Guo-JMIV-2013,kneip2011novel}.
In many situations, however, the internal camera calibration is unavailable, e.g. when working with crowd-sourced images. Consequently, methods have been proposed to jointly estimate the camera pose together with focal length~\cite{bujnak2008general,zheng2014general,wu2015p3,larsson2017making}. These methods have been extended to include also estimation of an unknown principal point~\cite{larsson2018camera}, and unknown radial distortion~\cite{josephson2009pose,larsson2017making}. The latter is particularly important for the wide-angle lenses commonly used in mobile phones and GoPro-style action cameras.
The absolute pose of fully uncalibrated perspective camera without radial distortion can be estimated from six point correspondence using the well-known DLT solution~\cite{DLT-1971}.
All these solutions assume a perspective camera model and are not suitable for cameras with rolling shutter (RS), unless the camera and the scene can be kept static.

Rolling shutter is omnipresent from consumer phones to professional SLR
cameras. Besides technical advantages, like higher frame-rate and longer exposure time per pixel, it is also cheaper to produce. The price to pay is that the rows of an ``image" are no longer captured synchronously, leading to motion-induced distortions and in general to a more complicated imaging geometry.

\begin{figure}[tb]
    \centering
    \includegraphics[trim={2cm 3cm 2cm 2cm},clip,width=0.32\columnwidth]{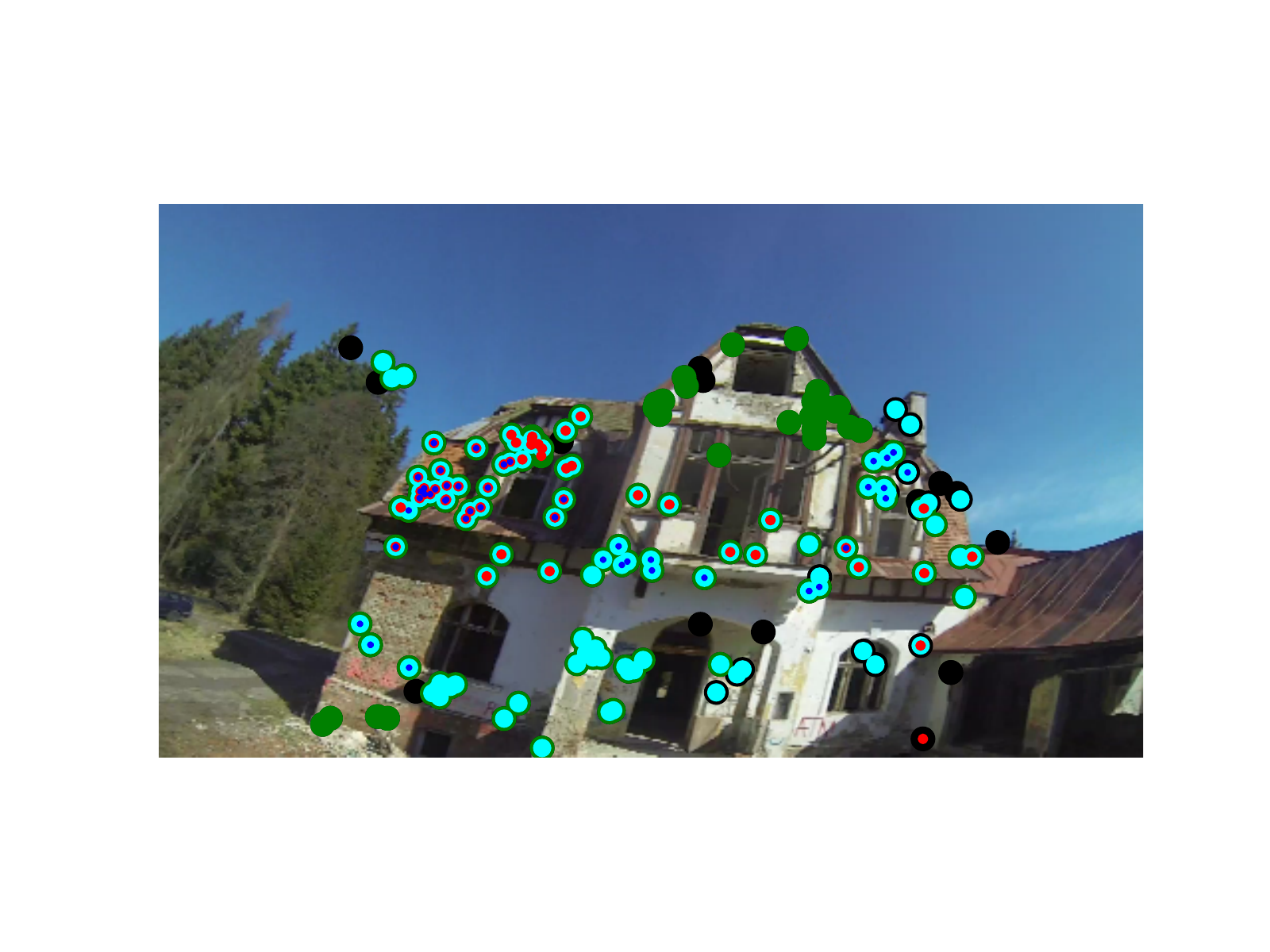}
    \includegraphics[trim={2cm 3cm 2cm 2cm},clip,width=0.32\columnwidth]{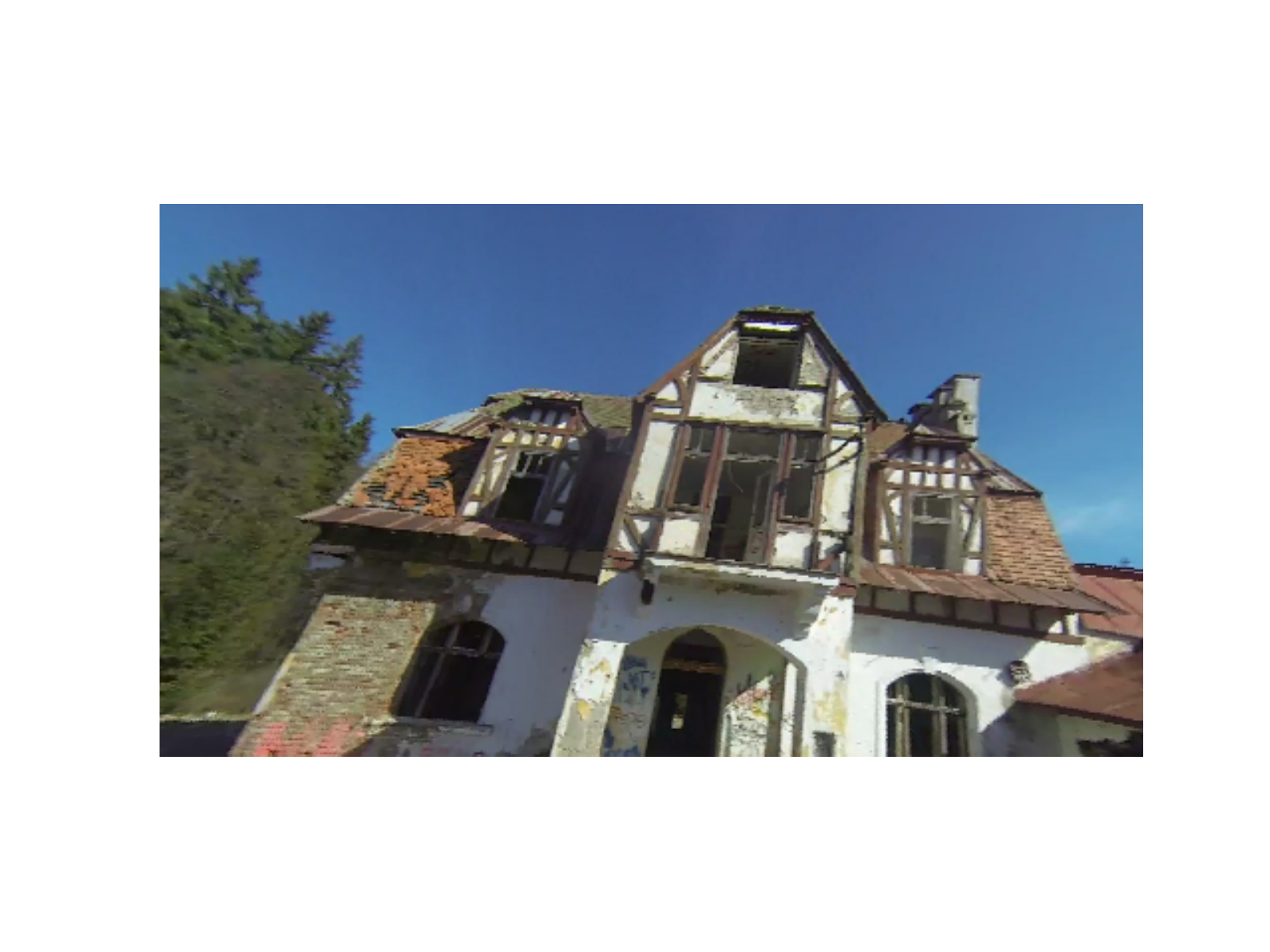}
    \includegraphics[trim={2cm 3cm 2cm 2cm},clip,width=0.32\columnwidth]{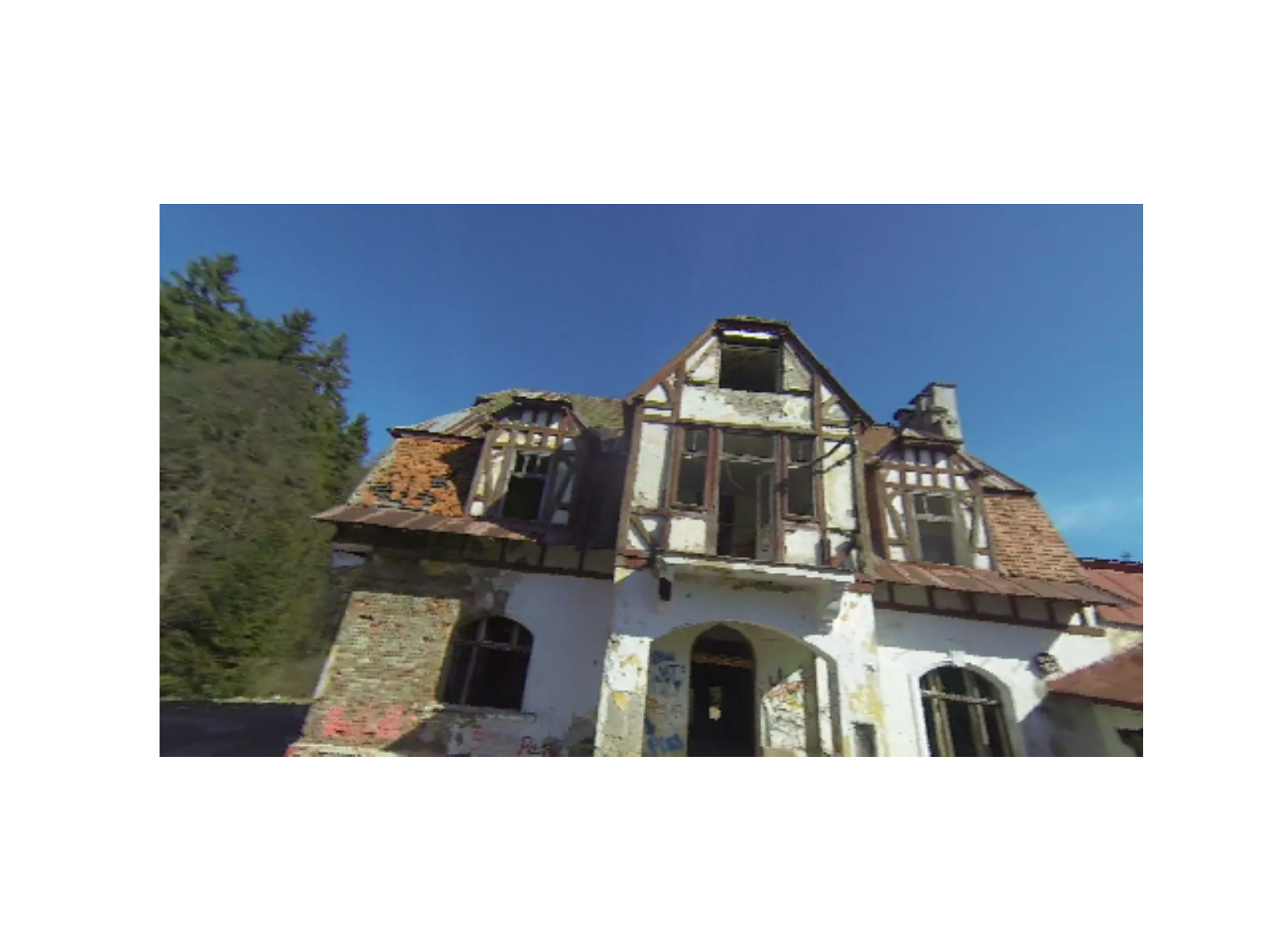}
    \caption{Removing RS and radial distortion simultaneously using our minimal absolute pose solver. The original image (left) with tentative correspondences (black) the inliers captured by P4Pfr+R6P~\cite{albl2019rolling} (blue) and subsequent local optimization (red) compared to inliers captured by the proposed R7Pfr (cyan) and subsequent local optimization (green). The correction using R7Pfr without local optimization (right) is better than by P4Pfr+R6P with local optimization (middle). }
    \label{fig:killer}
\end{figure}

\vspace{-2mm}
\paragraph{\bf Motivation:} While several minimal solutions have been proposed for the absolute pose of an RS camera with calibrated intrinsics~\cite{Albl-CVPR-2015,albl2019rolling,albl_rolling_2016,kukelova2018linear}, minimal solutions for uncalibrated RS cameras are missing.
One obvious way to circumvent that problem is to first estimate the intrinsic and radial distortion parameters while ignoring the rolling shutter effect, then recover the 3D pose and rolling shutter parameters with an absolute pose solver for calibrated RS cameras~\cite{Albl-CVPR-2015,albl2019rolling,kukelova2018linear}. Ignoring the deviation from the perspective projection in the first step can, however, lead to wrong estimates. For example, if the image point distribution is unfavourable, it may happen that RS distortion is compensated by an (incorrect) change of radial distortion, see Figure~\ref{fig:killer}.

\vspace{-2mm}
\paragraph{\bf Contribution:} We present the first minimal solutions for two rolling shutter absolute pose problems:
\begin{enumerate}
\vspace{-0.5em}
    \item absolute pose estimation of an RS camera with unknown focal length from 7 point correspondences; and
    \item absolute pose estimation of an RS camera with unknown focal length \emph{and} unknown radial distortion, also from 7 point correspondences.
\end{enumerate}
The new minimal solvers combine two ingredients: a recent, iterative approach introduced for pose estimation of calibrated RS cameras~\cite{kukelova2018linear}; and fast polynomial eigenvalue~\cite{PEP} and \gb solvers for comparatively simple, tractable systems of polynomial equations~\cite{cox2006using,larsson2017efficient}.
In experiments with synthetic and real data, we show that for uncalibrated RS cameras our new solvers find good estimates of camera pose, RS parameters,  focal length, and radial distortion. We demonstrate that the new all-in-one solvers outperform alternatives that sequentially estimate first perspective intrinsics, then RS correction and extrinsics.

\section{Related work}
The problem of estimating the absolute pose of a camera from a  minimal number of 2D-to-3D point correspondences is important in geometric computer vision. Minimal solvers are often the main building blocks for SfM~\cite{snavely2006photo,schoenberger2016CVPR} and localization pipelines~\cite{Sattler2017PAMI}. 
Therefore, during the last two decades a large variety of minimal absolute pose solvers for perspective cameras with or without radial distortion have been proposed. 

For estimating the absolute pose of a calibrated camera, three points are necessary and the resulting system of polynomial equations can be solved in a closed form~\cite{kneip2011novel}.
If the camera intrinsics and radial distortion are unknown, more point correspondences are required and the resulting systems of polynomial equations become more complex. The most common approach to solve such systems of polynomial equations is to use the \gb method~\cite{cox2006using} and automatic generators of efficient polynomial solvers~\cite{kukelova2008automatic,larsson2017efficient}. 

Most of the minimal absolute pose solvers have been developed using the \gb method. These include four or 3.5 point minimal solvers (P4Pf or P3.5Pf solvers) for the perspective camera with unknown focal length, and known or zero radial distortion~\cite{bujnak2008general,zheng2014general,wu2015p3,larsson2017making}, four point (P4Pfr) solvers for perspective cameras with unknown focal length and unknown radial distortion~\cite{josephson2009pose,bujnak2010new,larsson2017making}, and P4.5Pfuv solver for unknown focal length and unknown principal point~\cite{larsson2018camera}.





Recently, as RS cameras have become omnipresent, the focus has turned to problems of estimating the camera absolute pose from images containing RS effects. 
RS cameras motion models~\cite{Meingast2005} result in more complex systems of polynomial equations than perspective cameras models. Therefore, most of the existing RS absolute pose solvers use some model relaxations~\cite{Albl-CVPR-2015,albl2019rolling,kukelova2018linear}, scene assumptions such as planarity~\cite{Aitaider2006}, additional information e.g. from IMU~\cite{albl_rolling_2016} or a video sequence~\cite{Hedborg2011}, and a non-minimal number of point correspondences~\cite{Aitaider2006,Magerand2012}.
Moreover, all the existing solutions assume calibrated RS cameras, i.e., they assume that the camera intrinsic as well as radial distortion are known.

The first minimal solution to the absolute pose problem for a calibrated RS camera was presented in~\cite{Albl-CVPR-2015}. The proposed solver uses the minimal number of six 2D-to-3D point correspondences and the \gb method to generate an efficient solver. The proposed R6P is based on the constant linear and angular velocity model as in~\cite{Aitaider2006,Magerand2012,Hedborg2012}, but it uses the first order approximation to both the camera orientation and angular velocity, and, therefore, it requires an initialization of the camera orientation, e.g., from P3P~\cite{Fischler-Bolles-ACM-1981}.
It is shown in~\cite{Albl-CVPR-2015} that the proposed R6P solver significantly outperforms the P3P solver in terms of camera pose precision and the number of inliers captured in the RANSAC loop.
The R6P solver was extended in~\cite{albl2019rolling} by linearizing only the angular velocity and also by proposing a faster solution to the ``double-linearized'' model. The model that linearizes only the angular velocity does not require any initialization of the camera orientation, however it results in a slower and more complicated solver. Moreover, it is shown  in~\cite{albl2019rolling} that such solver usually produces similar results as the ``double-linearized'' solver initialized with 
P3P~\cite{Fischler-Bolles-ACM-1981,kneip2011novel}.

The double-linearized model~\cite{Albl-CVPR-2015,albl2019rolling} results in a quite complex system  of six quadratic equations in six unknowns with 20 solutions. The fastest solver to this problem presented in~\cite{albl2019rolling} runs $0.3 ms$ and is not suitable for real-time applications. 
Therefore, a further simplification of the double-linearized model was proposed~\cite{kukelova2018linear}.
The model in~\cite{kukelova2018linear} is based on the assumption that after the initialization with the P3P solver, the camera rotation is already close to the identity, and that in real applications, the rolling shutter rotation during the
capture is usually small. Therefore, some  nonlinear terms (monomials) in the double-linearized model are usually small,
sometimes even negligible.
Based on this assumption, a linear iterative algorithm was proposed in~\cite{kukelova2018linear}. 
In the first iteration, the algorithm substitutes negligible monomials with zeros. 
In each subsequent iteration, it substitutes these monomials with the estimates from the previous iteration. In this way, the original, complicated system of polynomial equations is approximated with a system of linear equations. This new linear iterative algorithm usually converges to the solutions of the original system in no more than five iterations and an order of magnitude faster than~\cite{albl2019rolling}.

%

Different from the above mentioned methods for calibrated RS cameras,  we combine the iterative scheme designed for calibrated RS cameras~\cite{kukelova2018linear} with fast generalized eigenvalue and \gb solvers~\cite{larsson2017efficient} for specific polynomial equation systems to solve the previously unsolved problem of estimating the absolute pose of an \emph{uncalibrated} RS camera (i.e., unknown RS parameters, focal length, and radial distortion) from a minimal number of point correspondences. 




\section{Problem formulation}
\label{sec:formulation}
For perspective cameras with radial distortion, the projection equation can be written as \begin{equation}
\alpha_i u(\V{x}_i,\V{\lambda}) = \M K [\, \M{R} \mid \M  C\,]\V{X}_i,
\label{eq:persp_proj}
\end{equation}
where $\M{R} \in SO(3)$ and $\V{C} \in \mathbb{R}^3$ are the rotation and translation bringing a 3D point $\V{X}_i= \left[x_i,y_i,z_i,1\right]^{\top}$ from the world coordinate system to the camera coordinate system,
$\V{x}_i = \left[r_i,c_i,1\right]^{\top}$ are the homogeneous coordinates of a measured distorted image point, $u\left(\cdot, \V{\lambda} \right)$ is an image undistortion function with parameters $\V{\lambda}$, and $\alpha_i \in \mathbb{R}$ is a scalar. 

Matrix $\M K$ is a $3 \times 3$ matrix known as the calibration matrix containing the intrinsic parameters of a camera.
%
Natural constraints satisfied by most consumer cameras with modern CCD or CMOS sensor are zero skew and the unit aspect ratio~\cite{HZ-2003}. 
The principal point~\cite{HZ-2003} is usually also close to the image center ($[p_x,p_y]^\top = [0,0]^\top$). Thus the majority of existing absolute pose solvers adhere to those assumptions, and we do so, too. Hence, we adopt calibration matrix 
\begin{equation}
\M K= {\rm diag}\left(f,f,1\right).
\label{eq:calib_f}
\end{equation}

For cameras with lens distortion, measured image coordinates $\V{x}_i$ have to be transformed into ``pinhole points'' with an undistortion function $u(\cdot, \V{\lambda})$.
For standard cameras, the radial component of the lens distortion is dominant, whereas the tangential component is negligible at this stage. Therefore, most camera models designed for minimal solvers consider only radial distortion\footnote{For maximum accuracy, tangential distortion can be estimated in a subsequent non-linear refinement.}. %
A widely used model represents radial lens distortion with a one-parameter division~\cite{fitzgibbon2001simultaneous}. This model is especially popular with absolute pose solvers thanks to its compactness and expressive power: it can capture even large distortions of wide-angle lenses (e.g., GoPro-type action cams) with a single parameter.  Assuming that the distortion center is in the image center, the division model is
\begin{equation}
   u\left(\V{x}_i,\lambda\right) =  u\left(\mat{l}{r_i\\c_i\\1},\lambda\right) = \mat{c}{r_i\\c_i\\1+\lambda(r_i^2+c_i^2)}.
   \label{eq:oneparam}
\end{equation}
Unlike perspective cameras, RS cameras capture every image row (or column) at a different time, and consequently, reveal the presence of relative motion between the camera and the scene at a different position.
Camera rotation $\M{R}$ and translation $\V{C}$ are, therefore, 
functions of the image row $r_i$ (or column). Together with the calibration matrix $\M{K}$ of~\eqref{eq:calib_f} and the distortion model~\eqref{eq:oneparam}, the projection equation of RS cameras is
\begin{equation}
\alpha_i u(\V{x}_i,\lambda) =
\alpha_i \mat{c}{r_i\\c_i\\1+\lambda(r_i^2+c_i^2)} =
\mat{ccc}{
f & 0 & 0 \\
0 & f & 0 \\
0 & 0 & 1
} [\M{R}(r_i)\mid \M  C(r_i)]\V{X}_i.
\label{eq:proj_rs}
\end{equation}
Let $\M{R}_0$ and $\M{C}_{0}$ be the unknown rotation and translation
of the camera at time $\tau= 0$, which denotes the acquisition time of the middle row $r_0 \in \mathbb{R}$. Then, for the short time-span required to record all rows of a frame (typically $<$ 50~ms), the translation $\V{C}(r_i)$ can be approximated by a constant velocity model~\cite{Albl-CVPR-2015,Saurer2013,Magerand2012,Meingast2005,Hedborg2012,Aitaider2006}:
\begin{equation}
\V{C}(r_i) = \V{C}_0 + (r_i-r_0)\V{t},
\label{eq:C_ri}
\end{equation}
with the translational velocity $\V{t}$. 

The rotation $\M{R}(r_i)$, on the other hand, can be decomposed into two parts: the initial orientation $\M{R}_0$ of $r_0$, and the change of the orientation relative to it: $\M{R}_{\V{w}}(r_i-r_0)$.
In~\cite{Magerand2012,Albl-CVPR-2015}, it was established that for realistic motions it is usually sufficient to linearize $\M{R}_{\V{w}}(r_i-r_0)$ around the initial rotation $\M{R}_0$ via the first-order Taylor expansion. Thereby the RS projection~\eqref{eq:proj_rs} becomes
\begin{equation}
\alpha_i \mat{c}{r_i\\c_i\\1+\lambda(r_i^2+c_i^2)} = \M{K}\left[ \left(\M{I}+(r_i-r_0)[\V{w}]_\times\right)\M{R}_0 \mid \V{C}_0+(r_i-r_0)\V{t}\right] \V{X}_i,
\label{eq:model_lin}
\end{equation}
%
where $[\V{w}]_\times$ is the skew-symmetric matrix for the vector $\V{w} = \left[w_1,w_2,w_3\right]^\top$.
%
%

The linearized model~\eqref{eq:model_lin} is sufficient for all scenarios except for the most extreme motions (which anyway present a problem due to motion blur that compromises keypoint extraction).

Unfortunately, the system of polynomial equation~\eqref{eq:model_lin} is rather complex even with the linearized rolling shutter rotation.  Already for calibrated RS camera and assuming Cayley parametrization of $\M{R}_0$, this model results in 
six equations of degree three in six unknowns and 64 solutions~\cite{albl2019rolling}.

Therefore, following~\cite{Albl-CVPR-2015,albl2019rolling}, we employ another linear approximation to the camera orientation $\M{R}_0$ to have the double-linearized model:
\begin{equation}
\alpha_i \mat{c}{r_i\\c_i\\1+\lambda(r_i^2+c_i^2)} = \M{K}\left[ \left(\M{I}+(r_i-r_0)[\V{w}]_\times\right)\left(\M{I}+[\V{v}]_\times\right) \mid \V{C}_0+(r_i-r_0)\V{t}\right] \V{X}_i.
\label{eq:model_double_lin}
\end{equation}
This model leads to a simpler way of solving the calibrated RS absolute pose from $\geq$ six 2D-3D point correspondences than the model in~\cite{albl2019rolling}.
However, the drawback of this further simplification is the fact that, other than the relative intra-frame rotation, the absolute rotation $\M{R}_0$ can be of arbitrary magnitude, and therefore far from the linear approximation.
A practical solution for calibrated cameras is to compute a rough approximate pose with a standard P3P solver~\cite{Fischler-Bolles-ACM-1981,kneip2011novel}, align the object coordinate system to it so that the remaining rotation is close enough to identity, and then run the full RS solver~\cite{Albl-CVPR-2015,albl2019rolling,kukelova2018linear}.

The double-linearized model~\eqref{eq:model_double_lin} is simpler than the original one~\eqref{eq:model_lin}, but still leads to a complex polynomial system (for calibrated RS cameras a system of six quadratic equations in six unknowns with up to 20 real solutions), and is rather slow for practical use.
%
%
Therefore, further simplification of the double-linearized model was proposed in~\cite{kukelova2018linear}. That model uses the fact that both the absolute rotation (after P3P initialisation) \emph{and} the rolling shutter rotation $\V{w}$ are small.
As a consequence~\cite{kukelova2018linear} assumes that the nonlinear term $[\V{w}]_\times[\V{{v}}]_\times$ in~\eqref{eq:model_double_lin} is sufficiently small (sometimes even negligible).
With this assumption, one can further linearize the nonlinear term $[\V{w}]_\times[\V{v}]_\times$ in~\eqref{eq:model_double_lin}  by approximating $[\V{v}]_\times$ with $[\V{\hat{v}}]_\times$, while keeping the remaining linear terms as they are; which leads to an efficient iterative solution of the original system: solve a resulting linearized system to estimate all unknowns including $[\V{v}]_\times$, and iterate with updated [$\V{\hat{v}}]_\times \leftarrow [\V{v}]_\times$ until convergence. 
As initial approximation one can set $[\V{\hat{v}}]_\times=\V{0}$.

Here we are interested in RS cameras with \emph{unknown} focal length and radial distortion. In that, setting~\eqref{eq:model_double_lin} leads to a much more complicated system of polynomial equations that exceeds the capabilities of existing algebraic methods such as \gbs~\cite{cox2006using,larsson2017efficient,kukelova2008automatic}.
We adopt a similar relaxation as in~\cite{kukelova2018linear} and linearize 
$[\V{w}]_\times [\V{v}]_\times$ by substituting with the preliminary value $[\V{v}]_\times\leftarrow[\V{\hat{v}}]_\times$.
Without loss of generality, let us assume that $r_0 = 0$. Then, the projection equation for this relaxed model is
\begin{equation}
\alpha_i \mat{c}{r_i\\c_i\\1+\lambda(r_i^2+c_i^2)} = 
\M{K}\left[ \M{I}+r_i[\V{w}]_\times + [\V{v}]_\times +  r_i[\V{w}]_\times[\V{\hat{v}}]_\times \mid \V{C}_0+r_i\V{t}\right] \V{X}_i.
\label{eq:model_double_lin2}
\end{equation}

\section{Minimal solvers}

To develop efficient minimal solvers for uncalibrated RS cameras, we advance the idea of the calibrated iterative RS solver of~\cite{kukelova2018linear} by combining it with a generalized eigenvalue and efficient \gb solvers for specific polynomial equation systems. 

We develop two new solvers. They both first pre-rotate the scene with a rotation estimated using efficient perspective absolute pose solvers for uncalibrated cameras, i.e. the P3.5Pf~\cite{larsson2017making} and P4Pfr~\cite{larsson2017making}/P5Pfr~\cite{kukelova2013real}.
Then, they iterate two steps: (i) solve the system of polynomial equations derived from~\eqref{eq:model_double_lin2}, with fixed preliminary $\V{\hat{v}}$. (ii) update $\V{\hat{v}}$ with the current estimates of the unknown parameters.
The iteration is initialised with $\V{\hat{v}}=\V{0}$.
A compact summary in the form of pseudo-code is given in Algorithm~\ref{alg:iterative}.
\begin{algorithm}[t]
\footnotesize
\caption{Iterative absolute pose solver for uncalibrated RS camera [with unknown radial distortion]}
\label{alg:iterative}
\def\converge{\textrm{\it converge}}
\def\notconverge{\textrm{\it notconverge}}
\def\skipped{\textrm{\it skipped}}
\def\false{\textrm{ FALSE}}
\def\true{\textrm{ TRUE}}
\def\solve{\textrm{ solve}}
\def\eq{\textrm{ Eq.}}
\def\nsol{\textrm{solutions of}}
\def\error{\textrm{Residual error of}}
\def\evaluated{\textrm{evaluated on}}
\def\or{\textrm{\bf or}}
\newcommand{\sgn}{\mathrm{sgn}}
\begin{algorithmic}
\REQUIRE $\V{x}_i,\V{X}_i$, $\left\{i=1,\ldots,7\right\}$, $k_{\max}$, $\epsilon_{\rm err}$
\ENSURE $\V v$, $\V C_0$, $\V w$, $\V t$, $f$, $\left[\lambda\right]$
\STATE $\V{v}^0 \leftarrow \V{0}$, 
$k \leftarrow 1$ 
 \STATE $\M{R}_{\rm GS}$, $\V{C}_{\rm GS}$, $f_{\rm GS}$ $\leftarrow$ P4Pf($\V x_i,\V{X}_j$)~\cite{larsson2017making}  
 \STATE [$\M{R}_{\rm GS}$, $\V{C}_{\rm GS}$, $f_{\rm GS}$, $\lambda_{\rm GS}$ $\leftarrow$ P4Pfr($\V x_i,\V{X}_j$)~\cite{larsson2017making}]  
  \STATE $\V{X}_i \leftarrow \M{R}_{\rm GS}\V{X}_i$
\WHILE{$k<k_{\max}$ 
}
    \STATE $\V{\hat v} \leftarrow \V v^{k-1}$  
    \STATE $err_{\max}^k \leftarrow \infty$
    \STATE $\left[\lambda_{\rm RS}\right]$, $\V v_{\rm RS}$, $\V C_{0{\rm RS}}$, $\V w_{\rm RS}$, $\V t_{\rm RS}$, $f_{\rm RS}$,  $\leftarrow \solve \eq$~\eqref{eq:model_double_lin2} 
    \FOR{$j =1$  \TO $\# \nsol \eq~\eqref{eq:model_double_lin2}$}
      \STATE $err_j \leftarrow\error \eq$~\eqref{eq:model_double_lin} $\evaluated \left\{ \V v_{{\rm RS}j}, \V C_{0{\rm RS}j}, \V w_{{\rm RS}j}, \V t_{{\rm RS}j}, f_{{\rm RS}j},\left[\lambda_{{\rm RS}j}\right]\right\}$
       \IF{$err_j<err_{\max}$}
          \STATE $\left\{ \V v^k, \V C^k, \V w^k, \V t^k, f^k,\left[\lambda^k\right]\right\} \leftarrow \left\{ \V v_{{\rm RS}j}, \V C_{0{\rm RS}j}, \V w_{{\rm RS}j}, \V t_{{\rm RS}j}, f_{{\rm RS}j},\left[\lambda_{{\rm RS}j}\right]\right\}$
          \STATE $err_{\max}^k \leftarrow err_j$
       \ENDIF
    \ENDFOR
    \IF{ $err_{\max}^{k} < \epsilon_{\rm err}$ $\or$  $\left(|err_{\max}^k - err_{\max}^{k-1}| < \epsilon_{\rm err}\;  \& \; k>1 \right)$ }
        \RETURN $\left\{ \V v^k, \V C^k, \V w^k, \V t^k, f^k,\left[\lambda^k\right]\right\}$
    \ENDIF
        \STATE $k \leftarrow k+1$
\ENDWHILE
\RETURN $\left\{ \V v^{k-1}, \V C^{k-1}, \V w^{k-1}, \V t^{k-1}, f^{k-1}, \left[\lambda^{k-1} \right]\right\}$
\end{algorithmic}
\label{alg:iter}
\end{algorithm}
Note that after solving the polynomial system~\eqref{eq:model_double_lin2}, we obtain, in general, more than one feasible solution (where ``feasible" means real and geometrically meaningful values, e.g., $f>0$). To identify the correct one among them, we evaluate the (normalized) residual error of the original equations~\eqref{eq:model_double_lin}, 
and choose the one with the smallest error.

The described computational scheme of Algorithm~\ref{alg:iterative} covers both the case where the radial distortion is known and only the pose, focal length and RS parameters must be found, and the case where also radial distortion is unknown.
In the following, we separately work out the R7Pf solver for known radial distortion and the R7Pfr solver for unknown radial distortion. Both the cases require seven point correspondences.

\subsection{R7Pf - RS absolute pose with unknown focal length}
\label{sec:R7Pf}
In the first solver, we assume that the camera has a negligible radial distortion (since known, non-zero distortion can be removed by warping the image point coordinates). 
This R7Pf solver follows the iterative procedure of Algorithm~\ref{alg:iterative}. What remains to be specified is how to efficiently solve the polynomial system~\eqref{eq:model_double_lin2} with $\lambda=0$.

The R7Pf solver first eliminates the scalar values $\alpha_i$ by left-multiplying equation~\eqref{eq:model_double_lin2} with the skew-symmetric matrix $\left[\V{x}_i\right]_\times$ for the vector $\V{x}_i=\mat{ccc}{\, r_i & c_i & 1\, }^\top$.
Since the projection equation~\eqref{eq:model_double_lin2} is defined only up to scale, we multiply the whole equation with $q ={1\over f}$ ($f\neq 0$), resulting in
\begin{equation}
 \mat{ccc}{
 0 & -1 & c_i \\
 1 & 0 & -r_i \\
 -c_i & r_i & 0
 }
 \mat{ccc}{
1 & 0 & 0 \\
0 & 1 & 0 \\
0 & 0 & q
}\left[ \M{I}+r_i[\V{w}]_\times + [\V{v}]_\times +  r_i[\V{w}]_\times[\V{\hat{v}}]_\times \mid \V{C}_0+r_i\V{t}\right] \V{X}_i =\V{0}.
\label{eq:model_double_lin3}
\end{equation}
\eqref{eq:model_double_lin3} has 13 degrees of freedom (corresponding to 13 unknowns): $\V{v},\V{w},\V{C}_0,\V{t}$, and $q ={1\over f}$.
Since each 2D--3D point correspondence gives two linearly independent equations (only two equations in~\eqref{eq:model_double_lin3} are linearly independent due to the singularity of the skew-symmetric matrix), we need $6\frac{1}{2}$ point correspondences for a minimal solution.

Since half-points for which only one coordinate is known normally do not occur, we present a 7-point solver and just drop out one of the constraints in computing the camera pose, RS parameters, and focal length. The dropped constraint can be further used to filter out geometrically incorrect solutions.



After eliminating the scalar values $\alpha_i$, the R7Pf solver starts with equations corresponding to the $3^{\rm rd}$ row of~\eqref{eq:model_double_lin3} for $i=1,\dots, 7$.
These equations are linear in ten unknowns and do not contain the unknown $q ={1\over f}$, indicating that they are independent of focal length.
Let us denote the elements of unknown vectors by $\V{v} = \left[v_1,v_2,v_3\right]^\top$, $\V{w} = \left[w_1,w_2,w_3\right]^\top$, $\V{C}_0 = \left[C_{0x},C_{0y},C_{0z}\right]^\top$, and $\V{t} = \left[t_x,t_y,t_z\right]^\top$.
Then, the equations corresponding to the $3^{\rm rd}$ row of~\eqref{eq:model_double_lin3} for $i=1,\dots, 7$ can be written as
\begin{equation}
    \M M \V{y} = \V{0}\quad\quad (i=1,\dots,7),
    \label{eq:Mx}
\end{equation}
where $\M M$ is a $7 \times 11$ coefficient matrix and $\V{y}$ is a $11 \times 1$ vector of monomials: $\V{y} = \left[ v_1, v_2, v_3, w_1, w_2, w_3, C_{0x}, C_{0y}, t_x, t_y, 1\right]^\top$.
For points in the general configuration, the matrix $\M M$ in~\eqref{eq:Mx} has a 4-dimensional null-space, so we can write the unknown vector $\V{y}$ as a linear combination of four $11\times 1$ basis vectors $\V{y}_1,\V{y}_2,\V{y}_3$, and $\V{y}_4$ of that  null-space:
\begin{equation}
    \V{y} = \beta_1\V{y}_1+\beta_2\V{y}_2+\beta_3\V{y}_3+\beta_4\V{y}_4,
    \label{eq:null-param}
\end{equation}
where $\beta_j\ (j =1,\dots,4)$ are new unknowns. One of these unknowns, e.g. $\beta_4$, can be eliminated (expressed as a linear combination of the remaining three unknowns $\beta_1,\beta_2,\beta_3$), using the constraint on the last element of $\V{y}$, which by construction is 1. 

In the next step, the parameterization~\eqref{eq:null-param} is substituted into the equations corresponding to the $1^{\rm st}$ (or $2^{\rm nd}$) row of~\eqref{eq:model_double_lin3} for $i=1,\dots, 6$. Note that here we use only six of seven available equations. The substitution results in six polynomial equations in six unknowns $\beta_1,\beta_2,\beta_3, C_{0z}, t_z, q$, and 10 monomials 
$m= \left[ \beta_1q, \beta_1, \beta_2q , \beta_2, \beta_3q , \beta_3, C_{0z}q, t_zq, q, 1\right]$.
This is a system of six quadratic equations in six unknowns, which could be solved using standard algebraic methods based on \gbs~\cite{Cox-Little-etal-05} and automatic \gb solver generators~\cite{larsson2017efficient,kukelova2008automatic}. However, in this specific case, it is more efficient to transform it to a generalized eigenvalue problem (GEP) of size $6\times 6$, by rewriting it as
\begin{equation}
    q\M A_1 \left[ \beta_1, \beta_2, \beta_3, C_{0z}, t_z, 1\right]^\top =\M A_0 \left[ \beta_1, \beta_2, \beta_3, C_{0z}, t_z, 1\right]^\top,
     \label{eq:GEP}
\end{equation}
where $\M A_0$ and $\M A_1$ are $6 \times 6$ coefficient matrices.
Equation~\eqref{eq:GEP} can be solved using standard efficient eigenvalue methods~\cite{PEP}. Alternatively, one can simplify even further by  eliminating monomials $C_{0z}q$ and $t_zq$, thereby also eliminating two unknowns $C_{0z}$ and $t_z$, and then solving a GEP of size $4 \times 4$. The remaining unknowns are obtained by a back-substitution into~\eqref{eq:null-param}.

\subsection{R7Pfr - RS absolute pose with unknown focal length and unknown radial distortion}
\label{sec:R7Pfr}

The R7Pfr solver finds the solution of the minimal problem with unknown absolute pose, RS parameters, focal length, and radial distortion. Compared to the first solver, there is one additional degree of freedom (14 unknowns in total); hence, we need seven 2D-to-3D point correspondences.
R7Pfr follows the same iterative approach. 

After eliminating the scalar values $\alpha_i$ by left-multiplying equation~(\ref{eq:model_double_lin2}) with the skew-symmetric matrix $\left[u(\V{x}_i,\lambda)\right]_\times$ for $u(\V{x}_i,\lambda)=\mat{ccc}{\, r_i, c_i,  1+\lambda(r_i^2+c_i^2)\, }^\top$, and multiplying the complete system with $q ={1\over f} (\neq 0)$, we obtain
\begin{equation}
 \mat{ccc}{
 0 & d_i & c_i \\
 d_i & 0 & -r_i \\
 -c_i & r_i & 0
 }
 \mat{ccc}{
1 & 0 & 0 \\
0 & 1 & 0 \\
0 & 0 & q
}\left[ \M{I}+r_i[\V{w}]_\times + [\V{v}]_\times +  r_i[\V{w}]_\times[\V{\hat{v}}]_\times \mid \V{C}_0+r_i\V{t}\right] \V{X}_i =\V{0},
\label{eq:model_double_lin_radial}
\end{equation}
with $d_i = 1+\lambda(r_i^2+c_i^2)$ for $i=1,\dots, 7$.
%
The polynomial system~\eqref{eq:model_double_lin_radial} is more complicated than that without radial distortion, but the third row remains unchanged, as it is independent not only of focal length, but also of radial distortion. 
We can therefore proceed in the same way: find the 4-dimensional null space of matrix $\M M$, eliminate $\beta_4$, and substitute the parametrization~\eqref{eq:null-param} back into the $1^{\rm st}$ (or $2^{\rm nd}$) row of~\eqref{eq:model_double_lin_radial}, to obtain a system of seven quadratic polynomial equations in seven unknowns $\beta_1,\beta_2,\beta_3, C_{0z}, t_z, q, \lambda$, and 14 monomials 
$m= \left[ \beta_1q, \beta_1\lambda, \beta_1, \beta_2q , \beta_2\lambda, \beta_2, \beta_3q , \beta_3\lambda, \beta_3, C_{0z}q, t_zq, q, \lambda, 1\right]$.

In the next step, we eliminate the monomials $C_{0z}q$ and $t_zq$ (and, consequently, also two unknowns $C_{0z}$ and $t_z$) by simple Gauss-Jordan elimination.
The resulting system of five quadratic equations in five unknowns $\beta_1,\beta_2,\beta_3, q, \lambda$, and 12 monomials has ten solutions. Different from the R7Pf case, this system does not allow a straight-forward transformation to a GEP. Instead, we solve it with the \gb method using the automatic solver generator~\cite{larsson2017efficient}.

To find a solver that is as efficient as possible, we follow the recent heuristic~\cite{Larsson-CVPR18}. We generate solvers for 1000 different candidate bases and select the most efficient one among them. The winning solver performs elimination on a $26 \times 36$ matrix (compared to a $36 \times 46$ matrix if using the standard basis and grevlex monomial ordering) and eigenvalue decomposition of a $10\times 10$  matrix. The remaining unknowns are again obtained by the back-substitution to~\eqref{eq:null-param}.

\section{Experiments}
\label{sec:experiments}
We evaluate the performance of both presented solvers on synthetic as well as various real datasets. The main strength of the presented R7Pf and R7Pfr solvers lies in the ability to handle uncalibrated data, which often occurs in the wild. 

%
\begin{table}[t]
    \centering
      \caption{Average numbers of inliers for different methods.}
    \label{tab:real_inliers}
    \begin{tabular}{c|c|c|c|c}
         Dataset & P4Pfr+R6P & P4Pfr+R6P+LO & {\bf P4Pfr+R7Pfr} & {\bf P4Pfr+R7Pfr+LO} \\ \hline
         Gopro drone 1 & 45 & 162 & \bf{170} & \bf{203} \\ \hline
         Gopro drone 2 & 124 & 130 & \bf{126} & \bf{131} \\ \hline
         Gopro rollerc. & 130 & \bf{137} & \bf{132} & \bf{137} \\ \hline
         Xiaomi wide & 58 & 66 & \bf{64} & \bf{67} \\ \hline
         & P4Pf+R6P & P4Pf+R6P+LO & {\bf P4Pf+R7Pf} & {\bf P4Pfr+R7Pf+LO} \\ \hline
         Xiaomi standard & \bf{72} & 95 & 44 & \bf{110} \\ \hline
    \end{tabular}
\vspace*{-\baselineskip}
\end{table}

\subsection{Data setup}
\textbf{Synthetic data:} For the synthetic experiments, we generate random sets of seven points in the cube with side one. We simulate a camera with 60 degrees FOV at random locations facing the center of the cube at a distance between one to four. We generate 1000 samples for each experiment, with 10 increment steps for the parameters that are being varied. We generate the camera motion using constant translational and rotational velocity model and the radial distortion using the one parameter division model. Note that even though our solvers are based on these models, they use approximations to the RS motion and therefore the data is never generated with identical model that is being solved for.

\noindent
\textbf{Real data:}
We use altogether five datasets. Three outdoor captured by Gopro cameras, two of which are downloaded from Youtube and one was proposed in~\cite{albl2019rolling}. Two contain drone footage and one a handheld recording of a rollercoaster ride. We have conducted an offline calibration of the internal parameters and lens distortion of Gopro Hero 3 Black used in dataset Gopro drone 1 from~\cite{albl2019rolling} using the Matlab Calibration Toolbox. 

To create the ground truth we undistorted images grabbed from the entire videos and used them all in an open source SfM pipeline COLMAP~\cite{schoenberger2016CVPR} to reconstruct a 3D model. Of course, the images containing significant RS distortion were not registered properly or not at all, but the scene was sufficiently reconstructed from the images where RS distortions were insignificant. We then selected the parts of trajectory which have not been reconstructed well and registered the 2D features in those images to the 3D points in the reconstructed scene. This was done for datasets Gopro Drone 2 and Gopro rollercoaster. For Gopro drone 1 we had DSLR images of the scene and we could reconstruct the 3D model using those, which led to much better data overall. 

Furthermore, we captured two dataset using the Xiaomi Mi 9 smartphone with both the standard and the wide FOV camera. The standard FOV camera contains virtually no radial distortion whereas the wide FOV camera has moderate radial distortion. We reconstructed the scene using static images from the standard camera and then registered sequences with moving camera to the reconstruction.
\begin{figure}[t]
    \centering
    \includegraphics[width=0.32\columnwidth]{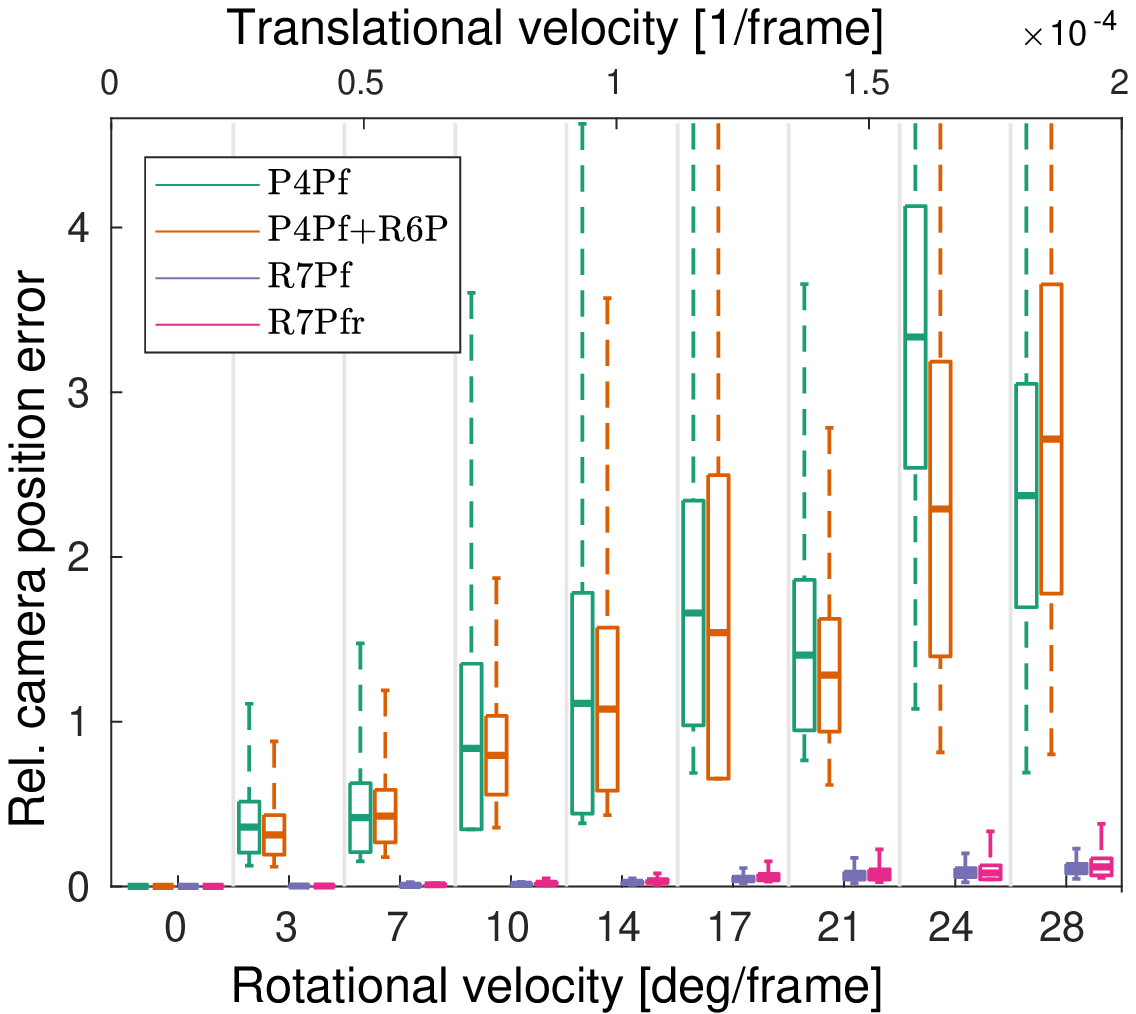}
    \includegraphics[width=0.32\columnwidth]{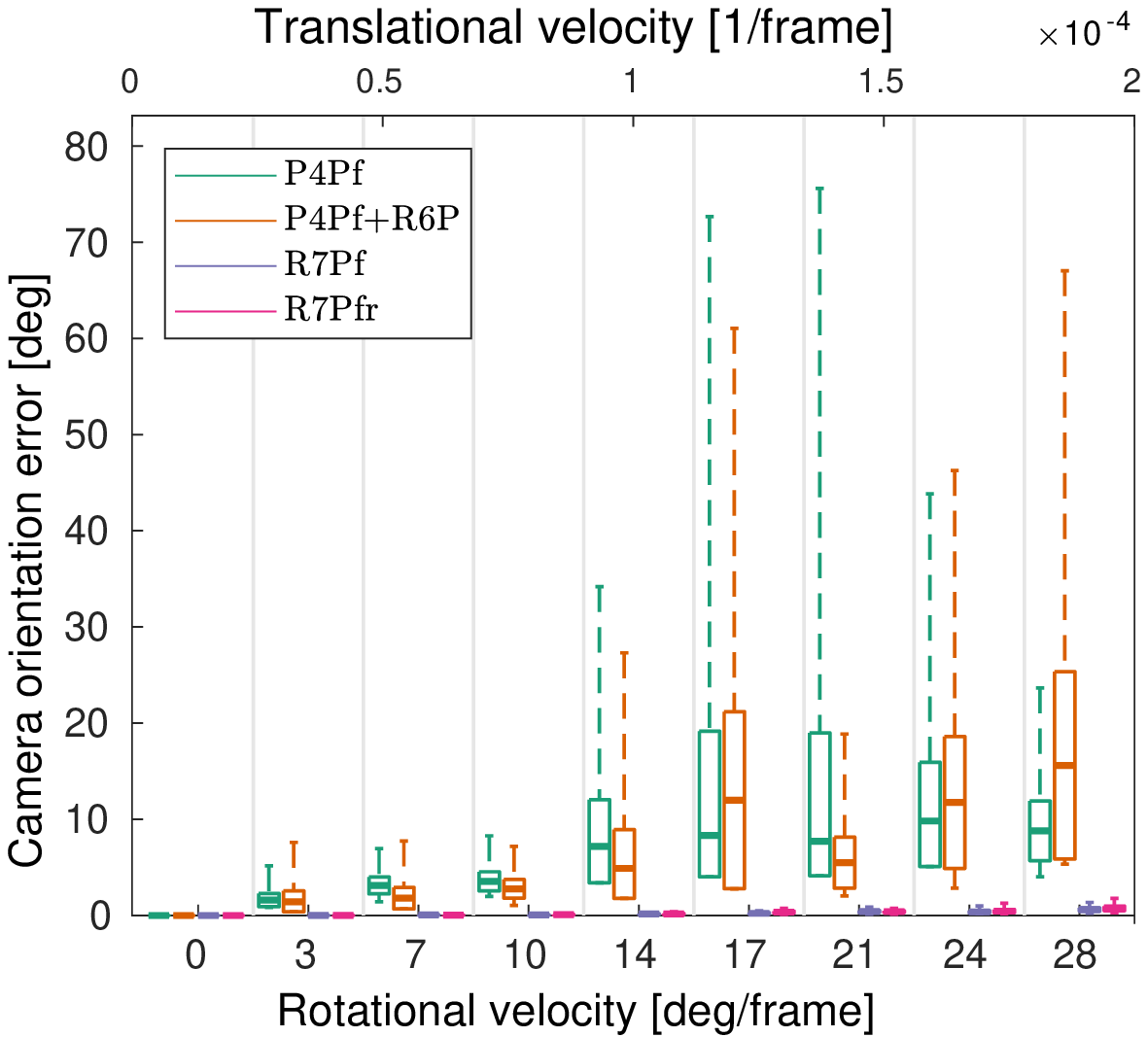}
    \includegraphics[width=0.32\columnwidth]{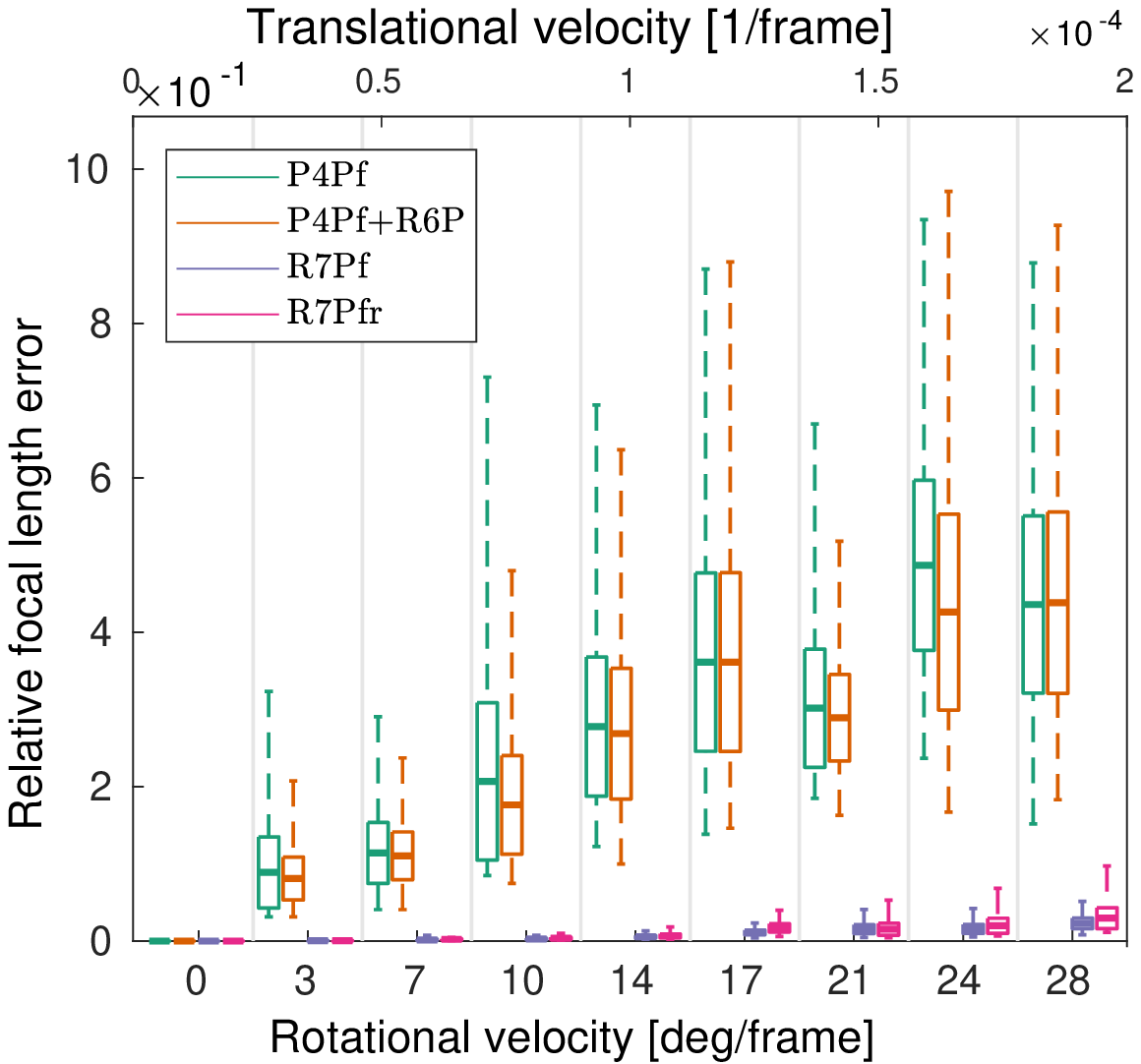}
    \caption{Increasing camera motion on synthetic data with unknown focal length. P4Pf struggles to estimate the correct pose and focal length in the presence of RS distortions whereas R7Pf (gray-blue) and R7Pfr (magenta) are able to cope with RS effects.}
    \label{fig:synth_f}
    \vspace*{-\baselineskip}
\end{figure}
\subsection{Compared methods}
\label{sec:exp_methods}

\begin{figure}[t]
    \centering
    \includegraphics[width=0.32\columnwidth]{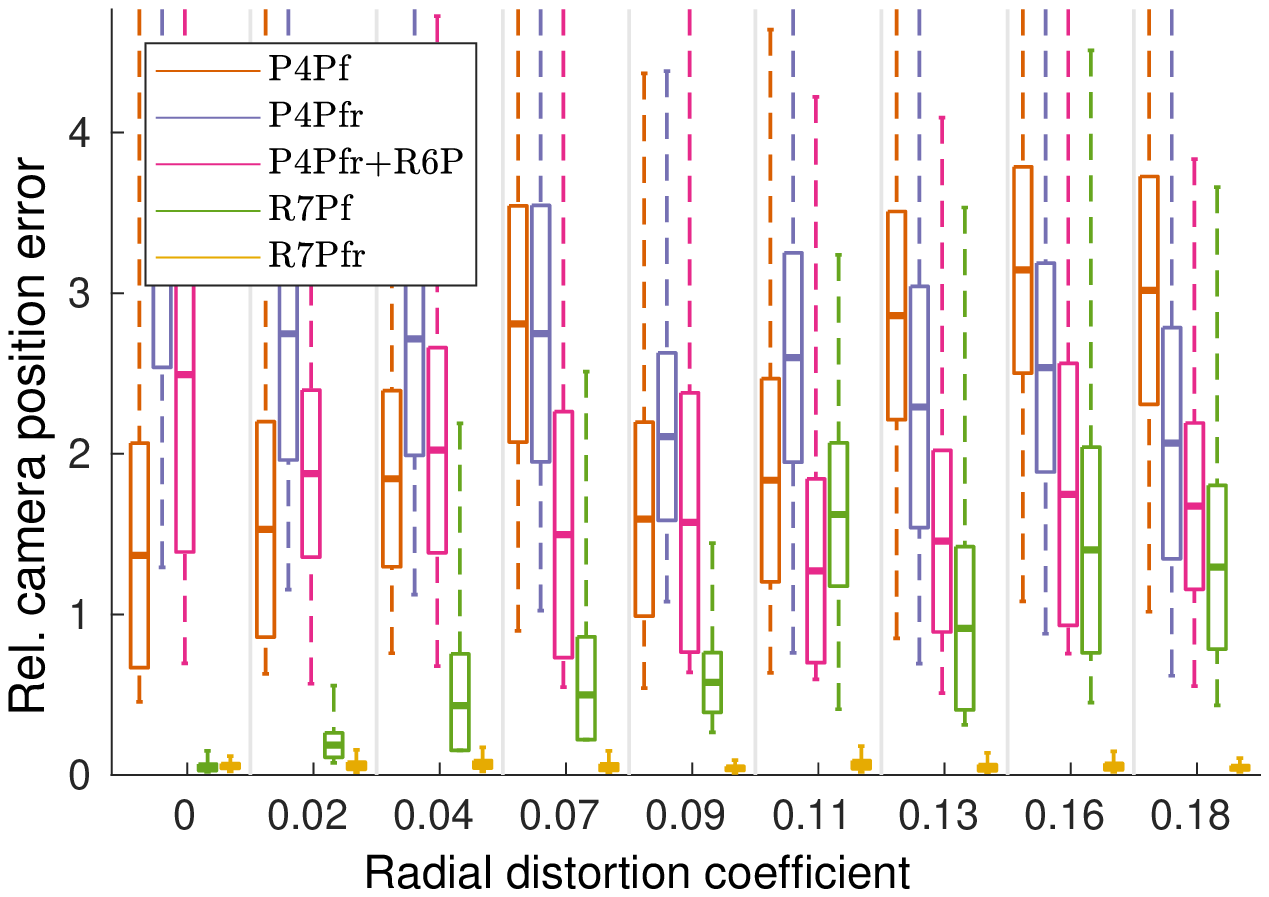}
    \includegraphics[width=0.32\columnwidth]{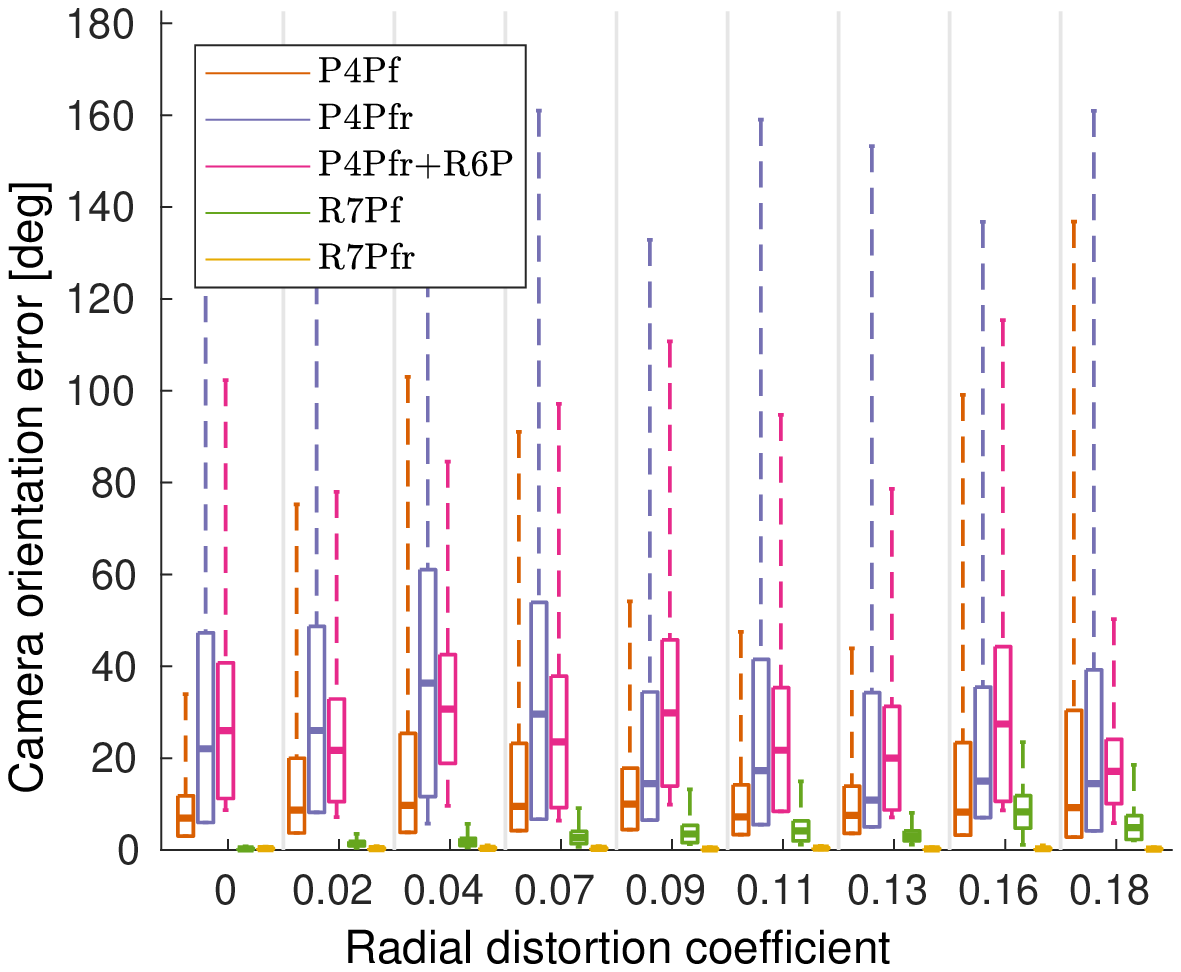}
    \includegraphics[width=0.32\columnwidth]{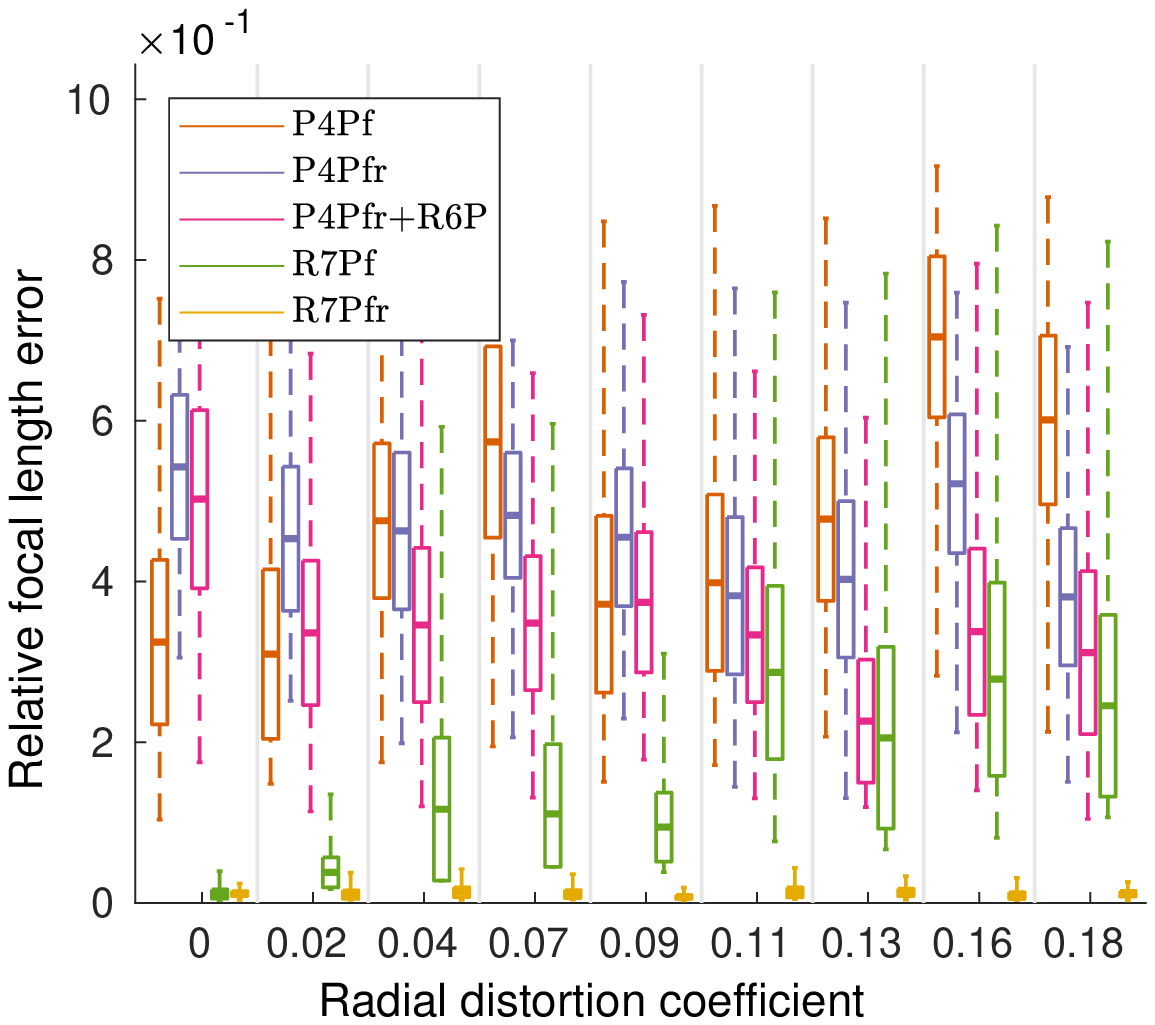}
    \caption{Increasing radial distortion and unknown focal length. RS motion is kept constant at a value of the middle of Figure~\ref{fig:synth_f}. P4Pf and P4Pfr are not able to estimate good pose and focal length under RS distortions, R7Pf slowly deteriorates with increasing radial distortion and R7Pfr performs well across the entire range.}
    \label{fig:synth_r}
\end{figure}

When neither the focal length nor the radial distortion coefficients are known, the state-of-the-art offers a 4-point solver to absolute pose, focal length (P4Pf)~\cite{larsson2017making} and radial distortion (P4Pfr)~\cite{larsson2017making}. In the presence of RS distortions, one can opt for the R6P algorithm~\cite{albl2019rolling} which, however, needs the camera calibration. We solve the problem simultaneously for both the RS parameters, focal length, and radial distortion, which until now could be emulated by running P4Pf or P4Pfr and subsequently R6P on the calibrated and/or undistorted data. That combination is the closest viable alternative to our method, so we consider it as the state-of-the-art and compare with it.

A common practice after robust estimation with a minimal solver and RANSAC is to polish the results with local optimization using all inliers~\cite{albl2019rolling}. 
An important question is whether a simpler model, in our case the baseline P4Pf/P4Pfr followed by R6P could be enough to initialize that local optimization and reach the performance of the direct solution with a more complex model, i.e., the proposed R7Pf/R7Pfr. In our experiments we evaluate also the non-linear optimization initialized by RANSAC and see if our solvers outperform the non-linear optimization of the baseline approach. 
\subsection{Evaluation metrics}
\label{sec:exp_metrics}
We use various metrics to compare with the state-of-the-art and to show the benefits of our solvers. A common practice~\cite{Albl-CVPR-2015,albl_rolling_2016,kukelova2018linear} is to use the number of inliers identified by RANSAC as the criterion to demonstrate the performance of minimal solvers on real data. We compare against both the state-of-the-art RANSAC output and the polished result after local optimization. 

To highlight the accuracy of the estimated radial distortion and the RS parameters, we use them to remove the radial distortion and the rotational rolling shutter distortion from the images. 

Due to the lack of good ground-truth for the camera poses, we evaluate them in two ways. First, we move the camera in place inducing only rotations, which resembles, e.g., an augmented or head-tracking scenario. In this case, the computed camera centers are expected to be almost static and we can show the standard deviation from the mean as a measure of the estimated pose error. Second, we evaluate qualitatively the case where the camera moves along a smooth trajectory.

\subsection{Results}
The experiments on synthetic data verify that the proposed solvers are able to handle unknown focal length, radial distortion and RS distortions. Figure~\ref{fig:synth_f} shows results on data with unknown focal length and increasing camera motion. The state-of-the-art P4Pf solver struggles to estimate the camera pose and the focal length accurately as the RS camera rotational and translational velocity increases, resulting in mean orientation errors up to 15 degrees and relative focal length error of 40\% when the motion is strongest. Given such poor initial focal length estimate from P4Pfr, R6P is not able to recover the pose any better. In contrast, both R7Pf and R7Pfr are able to estimate the pose and focal length accurately, keeping the mean rotation error under 1.0 degree and the relative focal length estimate error under 3\% even for the strongest motions. 

Next we evaluate the effect of increasing radial distortion and the performance of our R7Pfr solver, see Figure~\ref{fig:synth_r}. The magnitude of the RS motion is kept constant through the experiment at the value of about the middle of the previous experiment. First thing to notice is that P4Pfr is less stable under RS distortion, providing worse estimates than P4Pf when radial distortion is close to zero. As the radial distortion increases the performance of P4Pf becomes gradually worse and is outperformed by P4Pfr in the end. R6P initialized by P4Pfr is not able to improve the poor results of P4Pfr. R7Pf slowly deteriorates with increasing radial distortion and R7Pfr provides good results under all conditions. 

\begin{figure}[t]
    \centering
    \includegraphics[width=0.3\columnwidth]{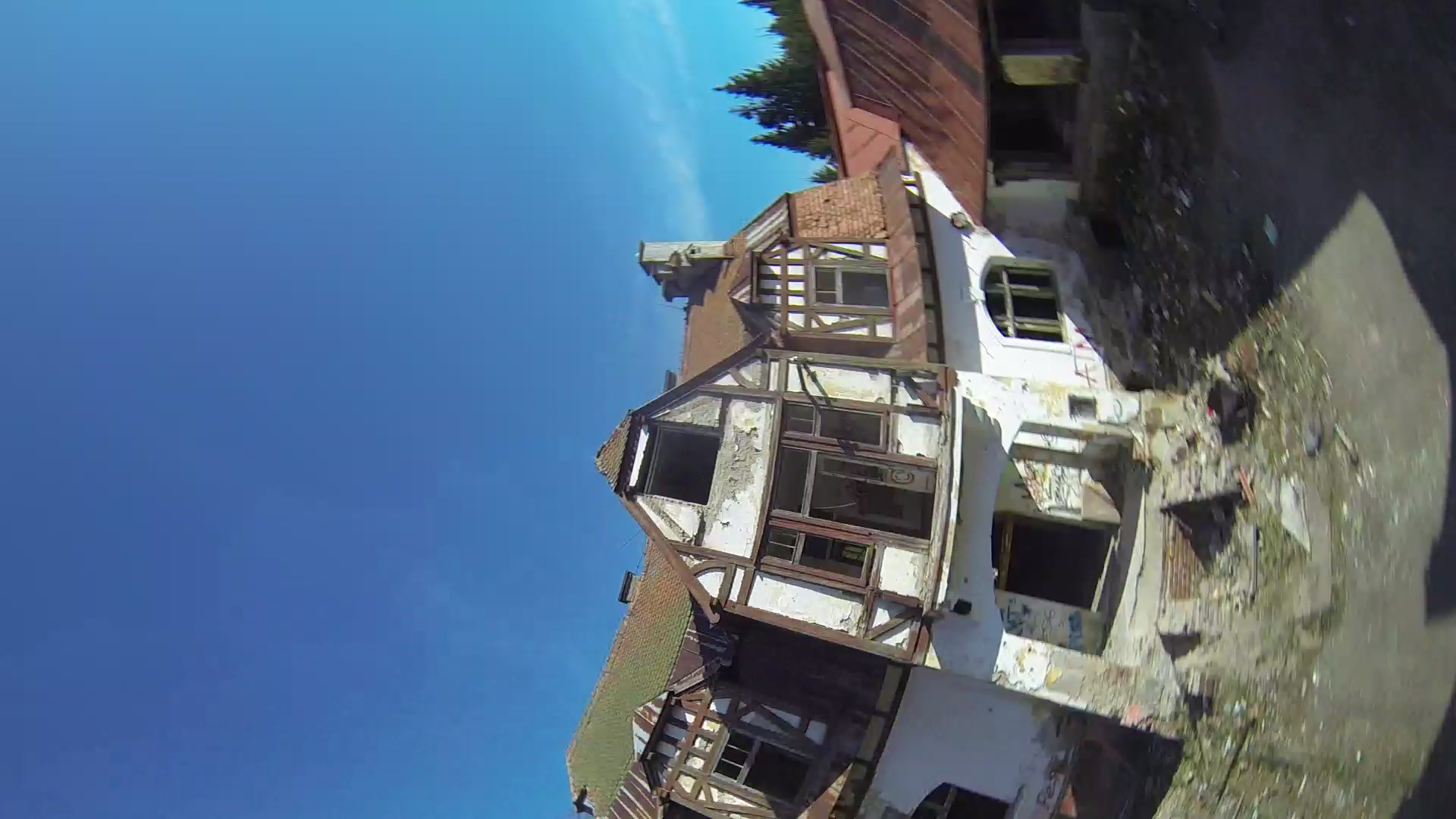}
    \includegraphics[width=0.3\columnwidth]{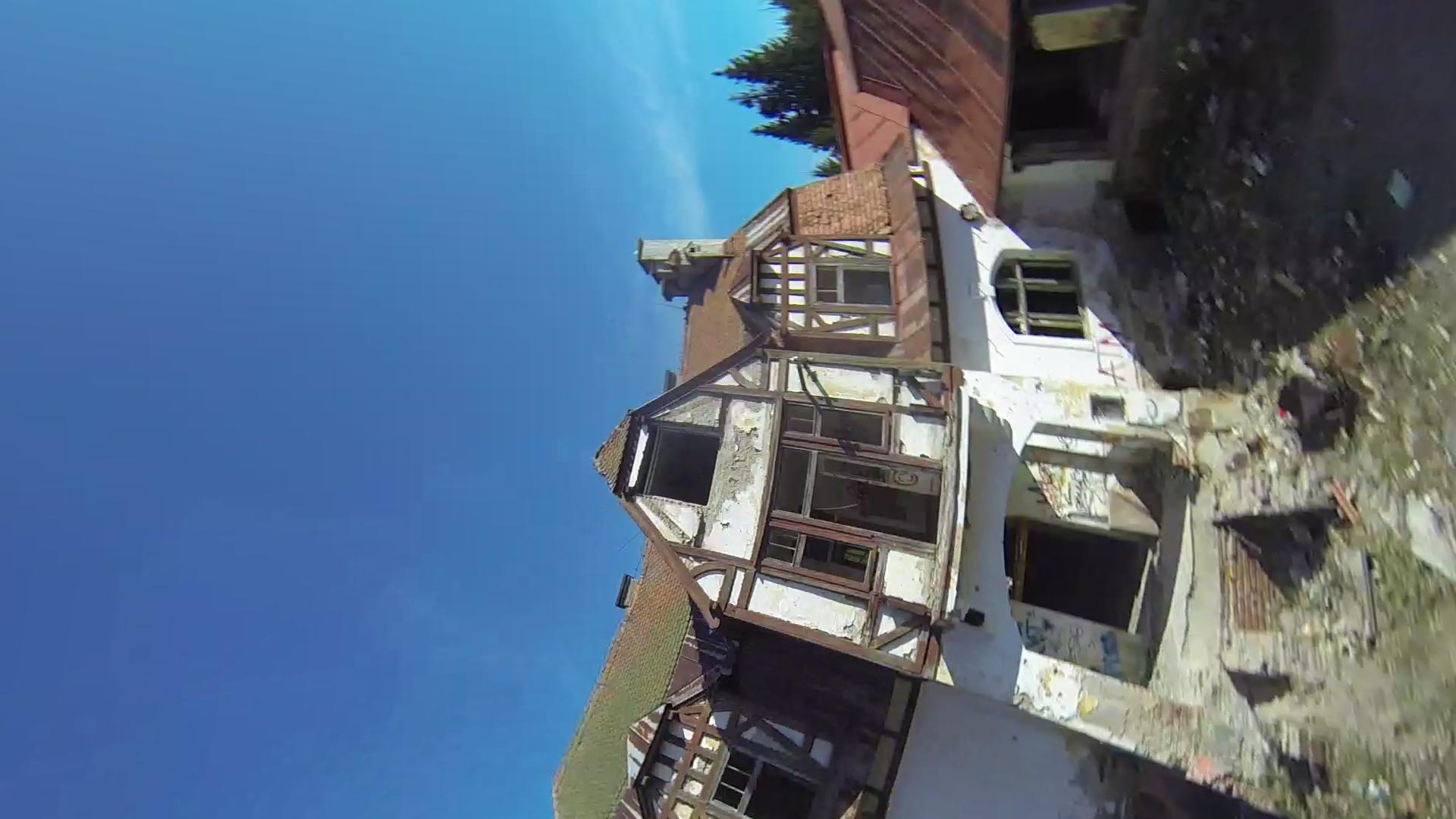}
    \includegraphics[trim={2cm 2.5cm 2cm 2.5cm},clip,width=0.3\columnwidth]{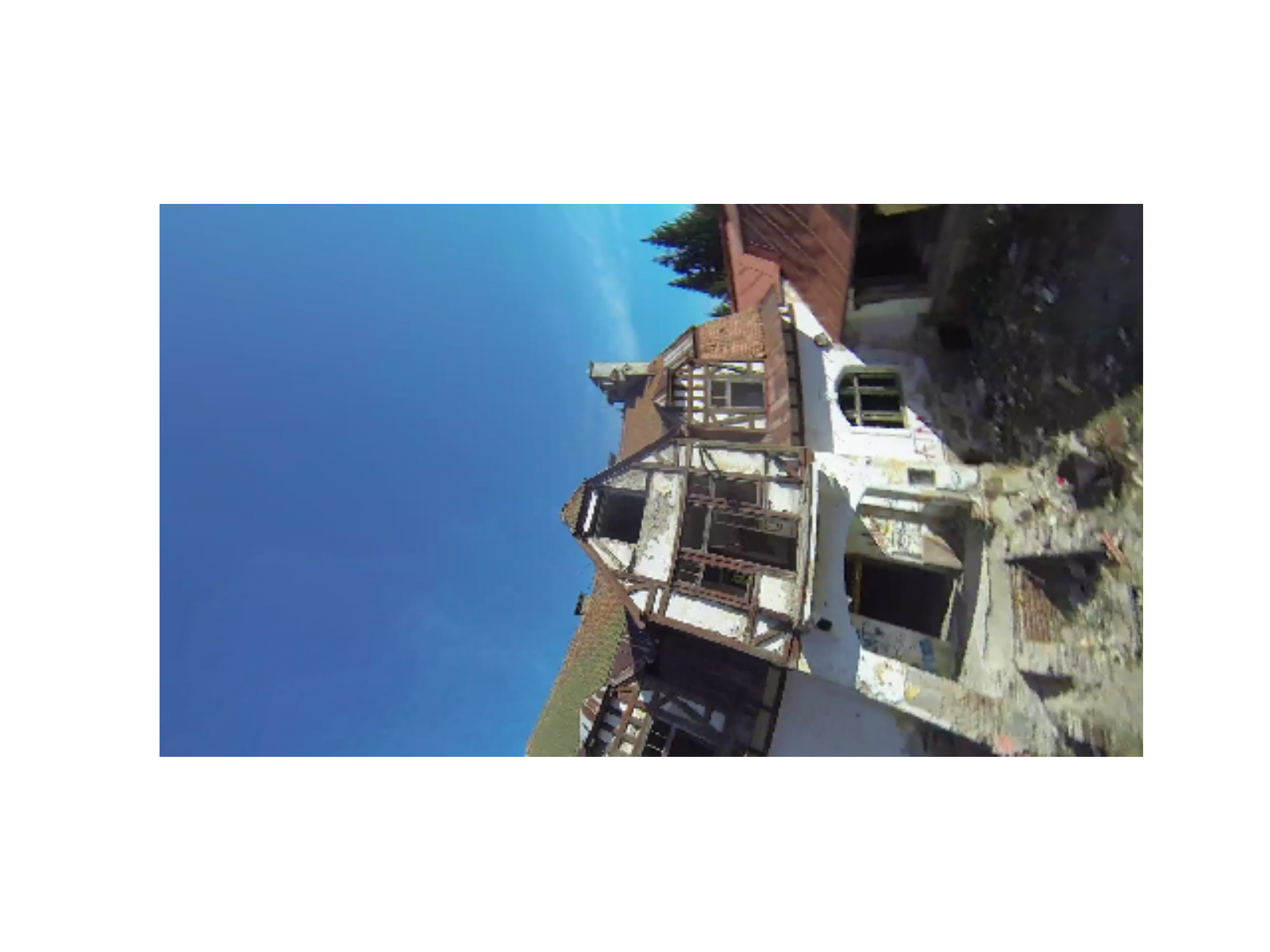}
    \caption{(left) Original image distorted by radial and RS distortion. (middle) {\bf Image undistorted by our R7Pfr}. (right) Image undistortion by a 3-parametric model estimated by Matlab Calibration Toolbox using a calibration board. We achieve comparable results as a method based on a calibration board.}
    \label{fig:nimrod_undist}
    \vspace*{-\baselineskip}
\end{figure}

\begin{table}[t]
    \centering
        \caption{Standard deviations from mean position of camera centers. The camera was purely rotating in these datasets; lower deviations mean more precise camera poses.}
    \begin{tabular}{c|c|c|c|c}
    Dataset & P4Pfr+R6P & P4Pfr+R6P+LO & P4Pfr+R7Pfr & P4Pfr+R7Pfr+LO \\ \hline
    Xiaomi wide & 25 & 39 & 20 & 20 \\
    \hline
    Xiaomi standard & 12 & 14 & 14 & 10 \\
    \end{tabular}
    \label{tab:camera_centers}
\end{table}

\begin{figure}[t]
    \centering
    \includegraphics[trim={2cm 3cm 2cm 2cm},clip,width=0.30\columnwidth]{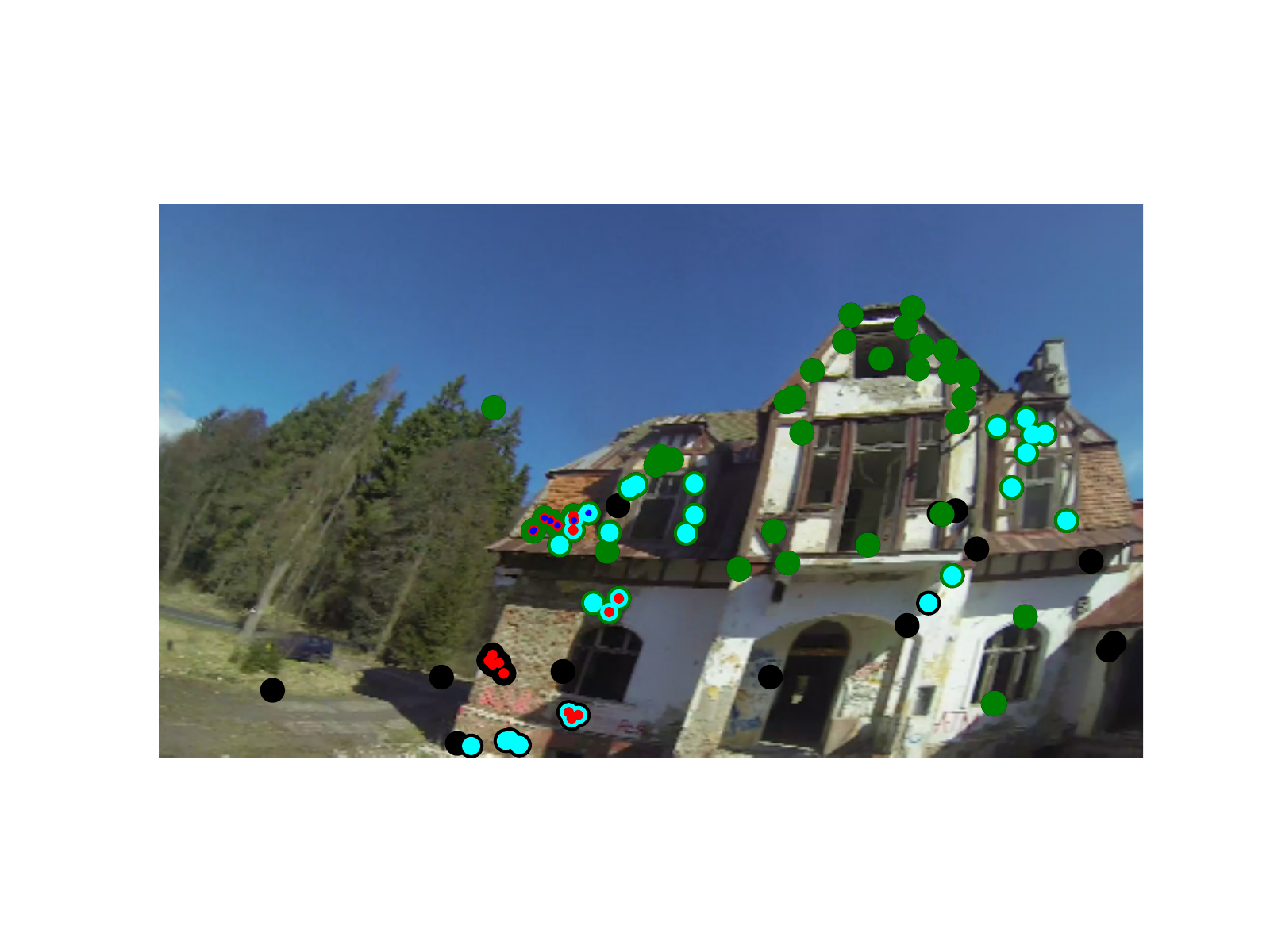}
    \includegraphics[trim={2cm 3cm 2cm 2cm},clip,width=0.30\columnwidth]{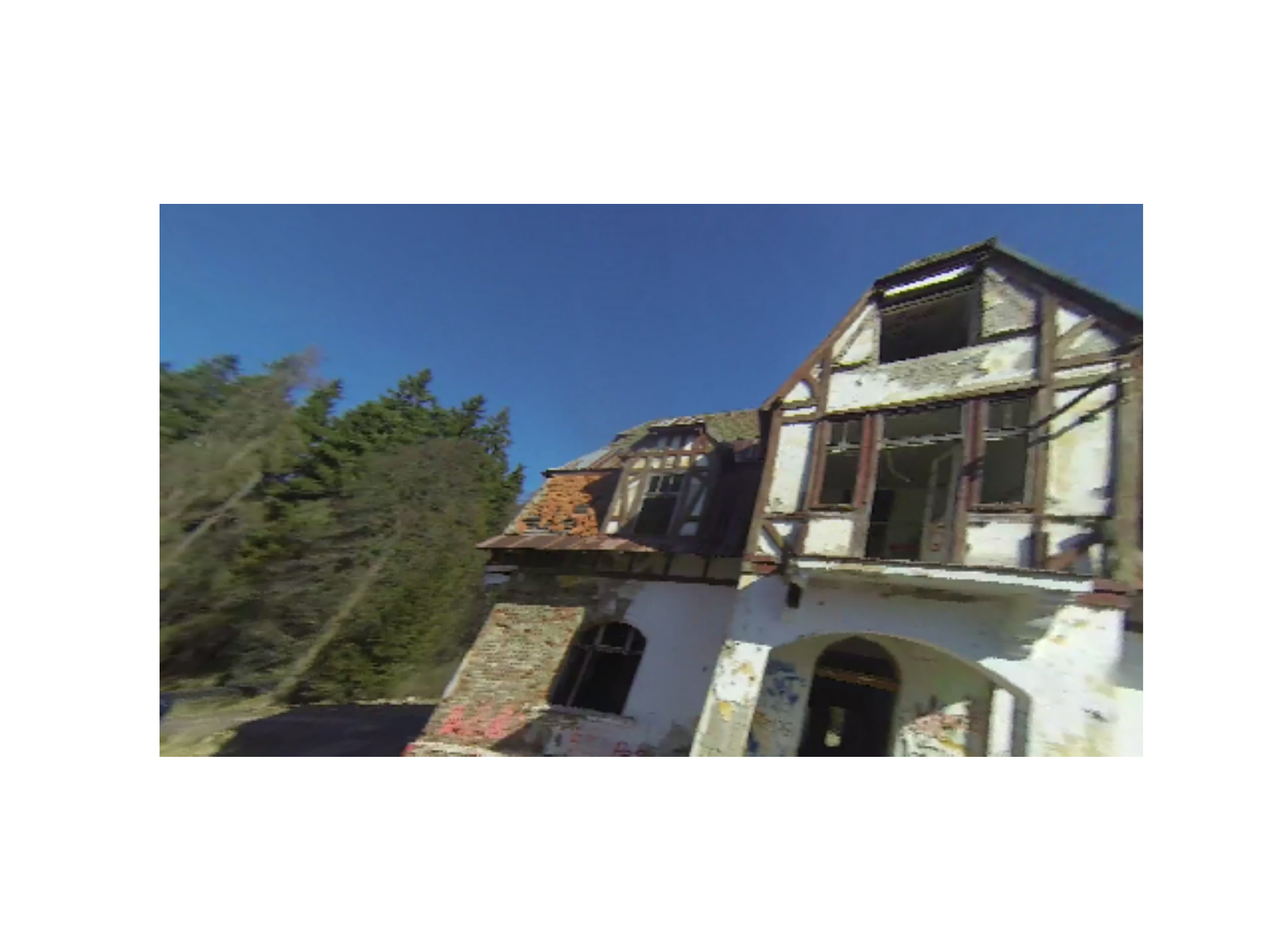}
    \includegraphics[trim={2cm 3cm 2cm 2cm},clip,width=0.30\columnwidth]{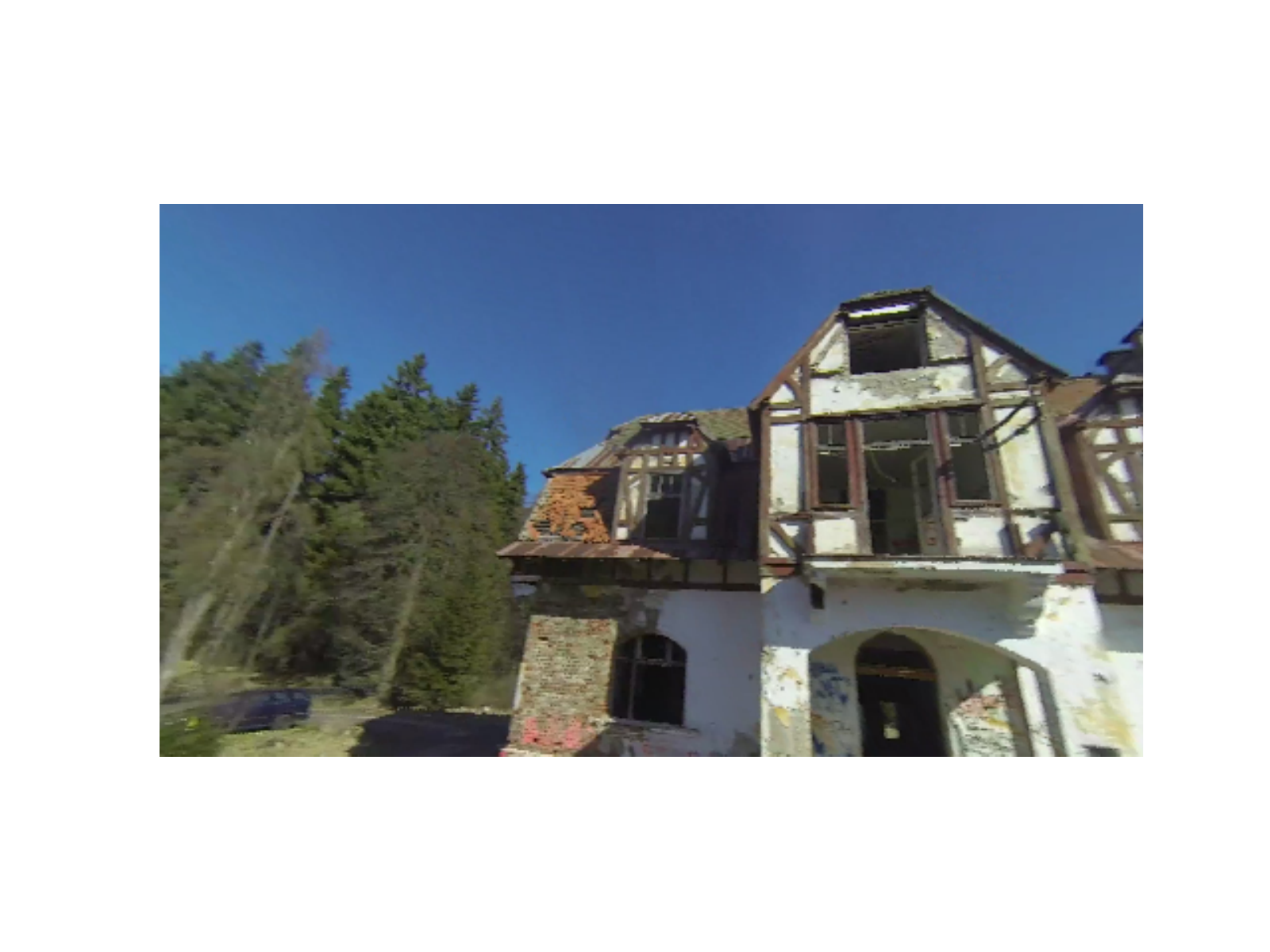}\\
    \includegraphics[trim={2cm 3cm 2cm 2cm},clip,width=0.30\columnwidth]{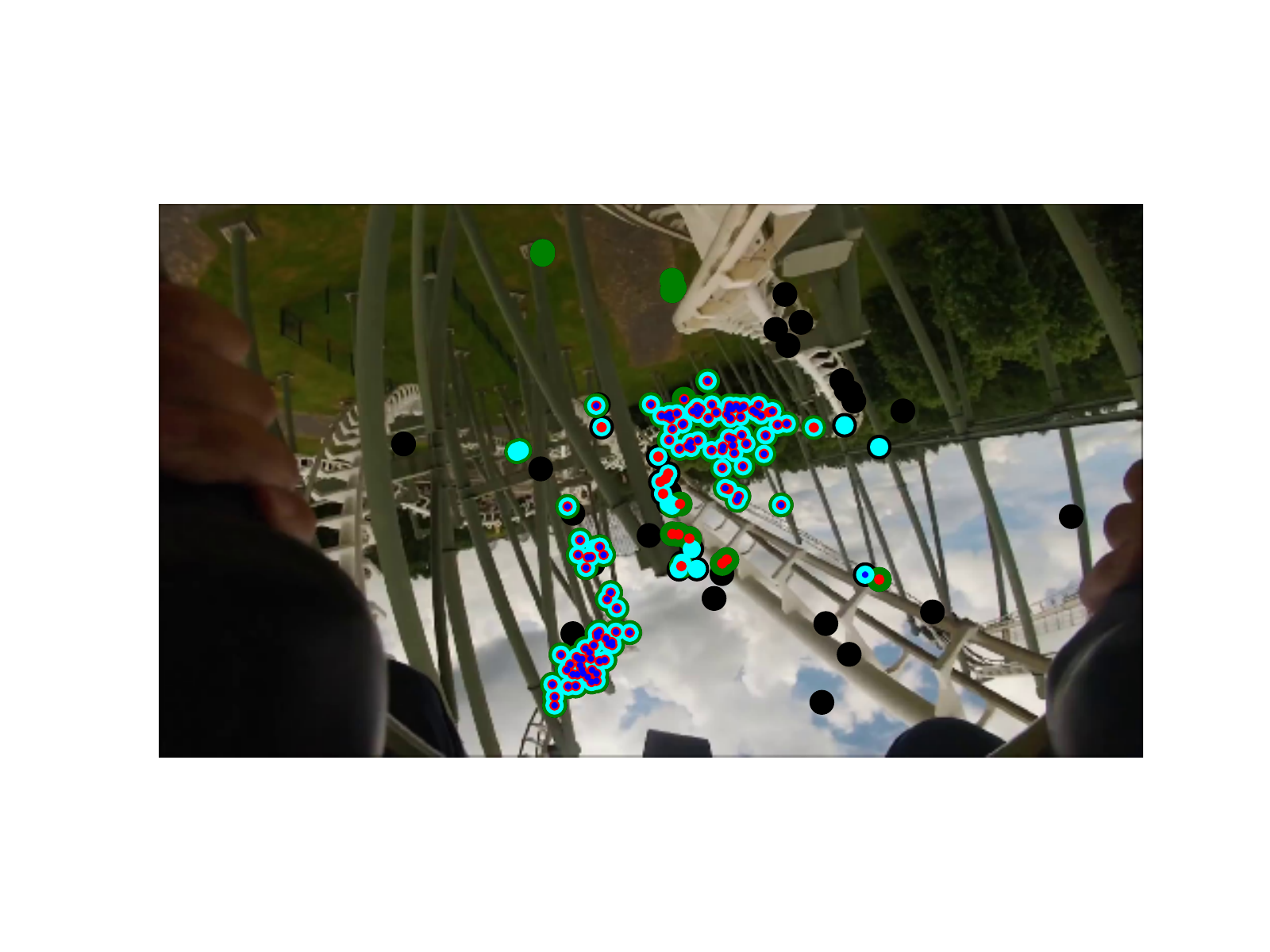}
    \includegraphics[trim={2cm 3cm 2cm 2cm},clip,width=0.30\columnwidth]{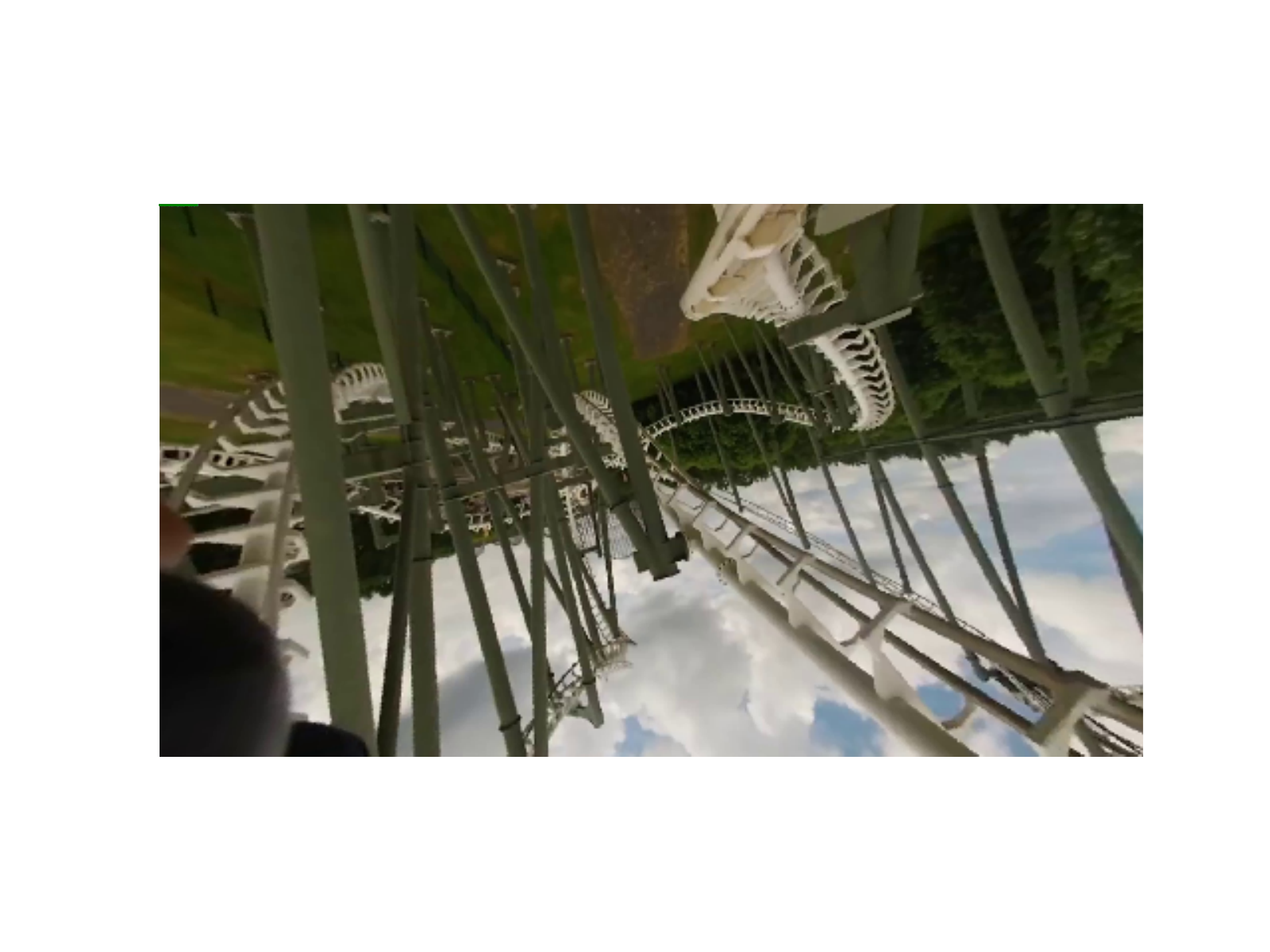}
    \includegraphics[trim={2cm 3cm 2cm 2cm},clip,width=0.30\columnwidth]{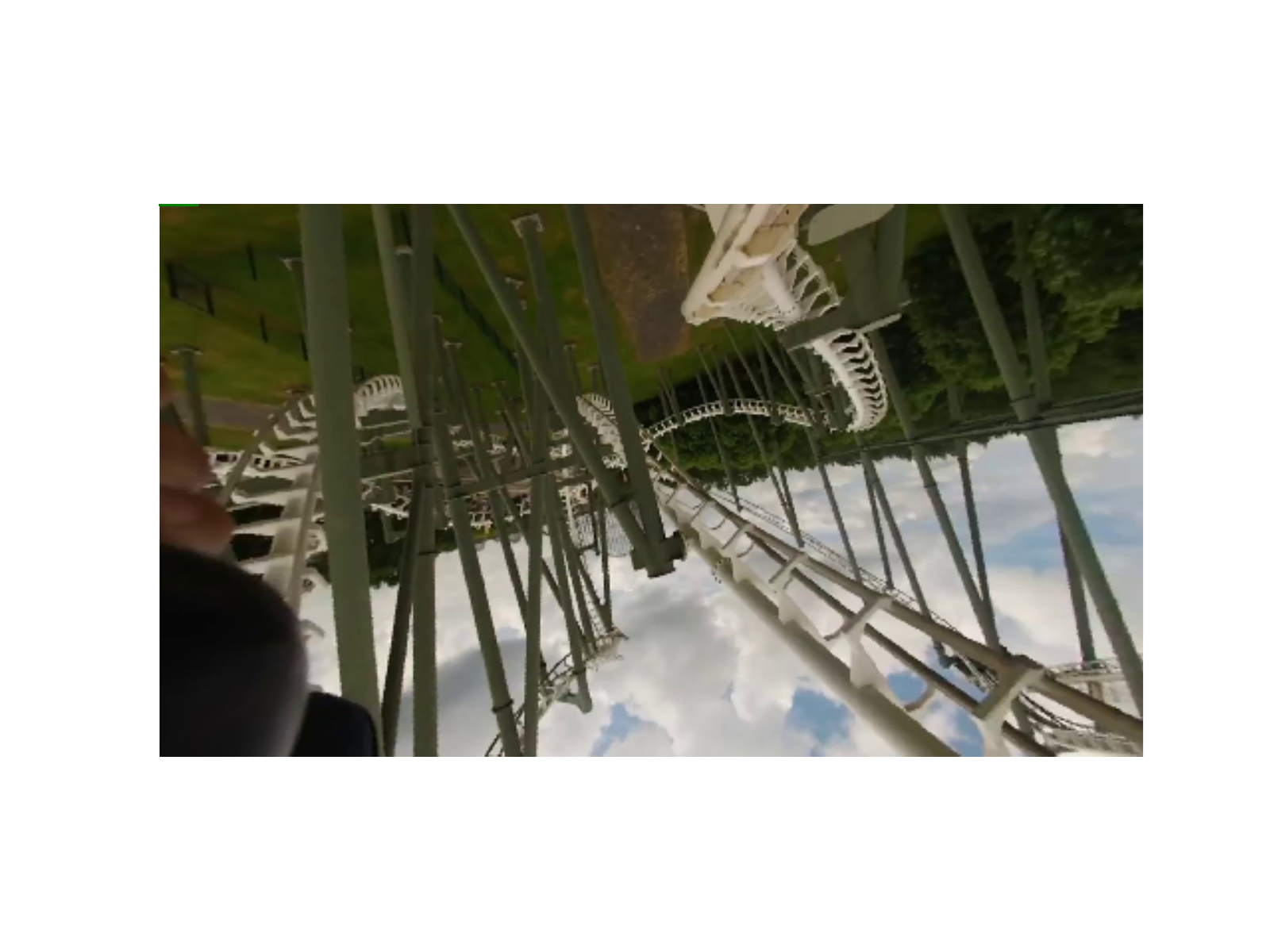}\\
    \includegraphics[trim={2cm 3cm 2cm 2cm},clip,width=0.30\columnwidth]{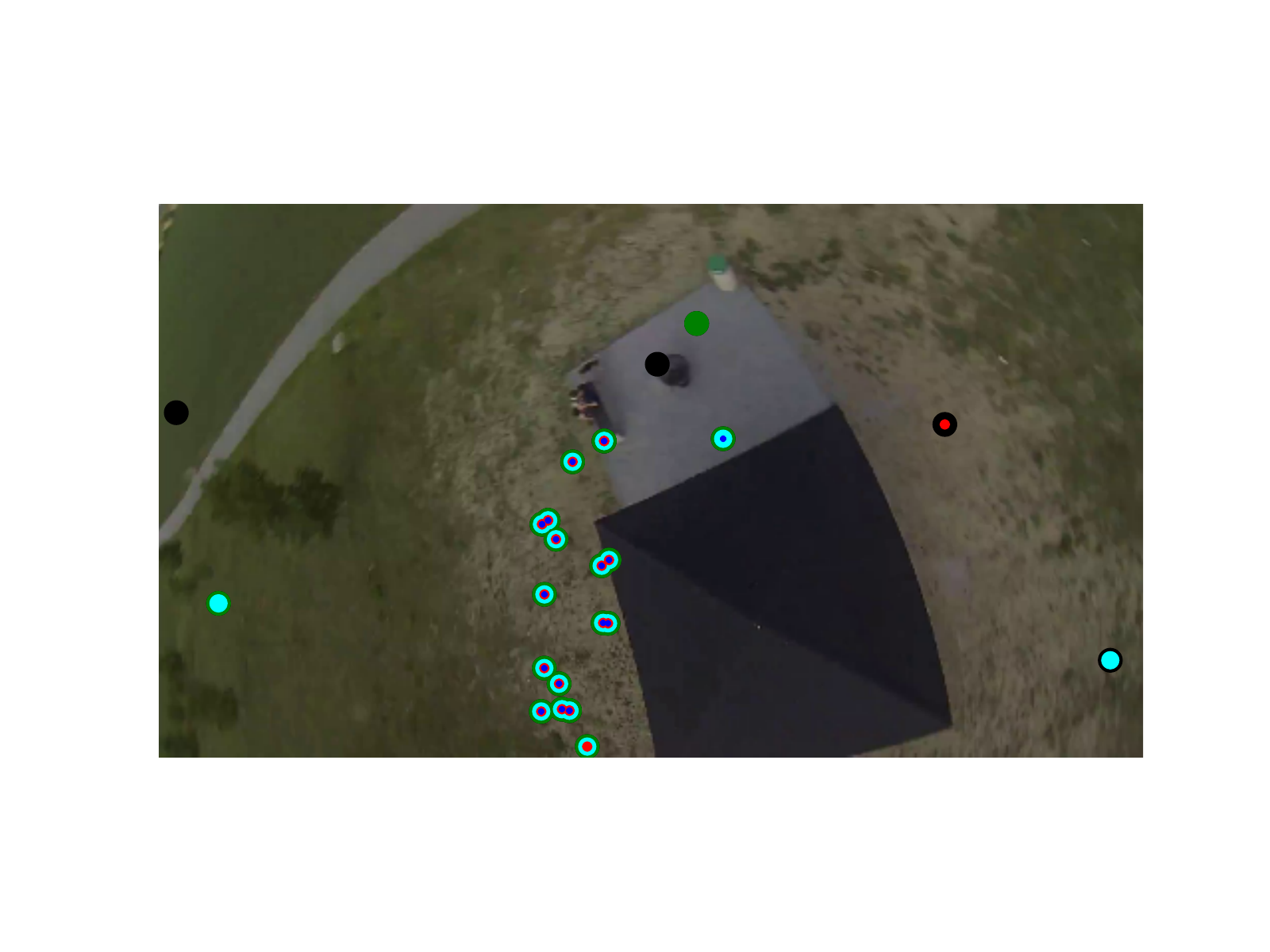}
    \includegraphics[trim={2cm 3cm 2cm 2cm},clip,width=0.30\columnwidth]{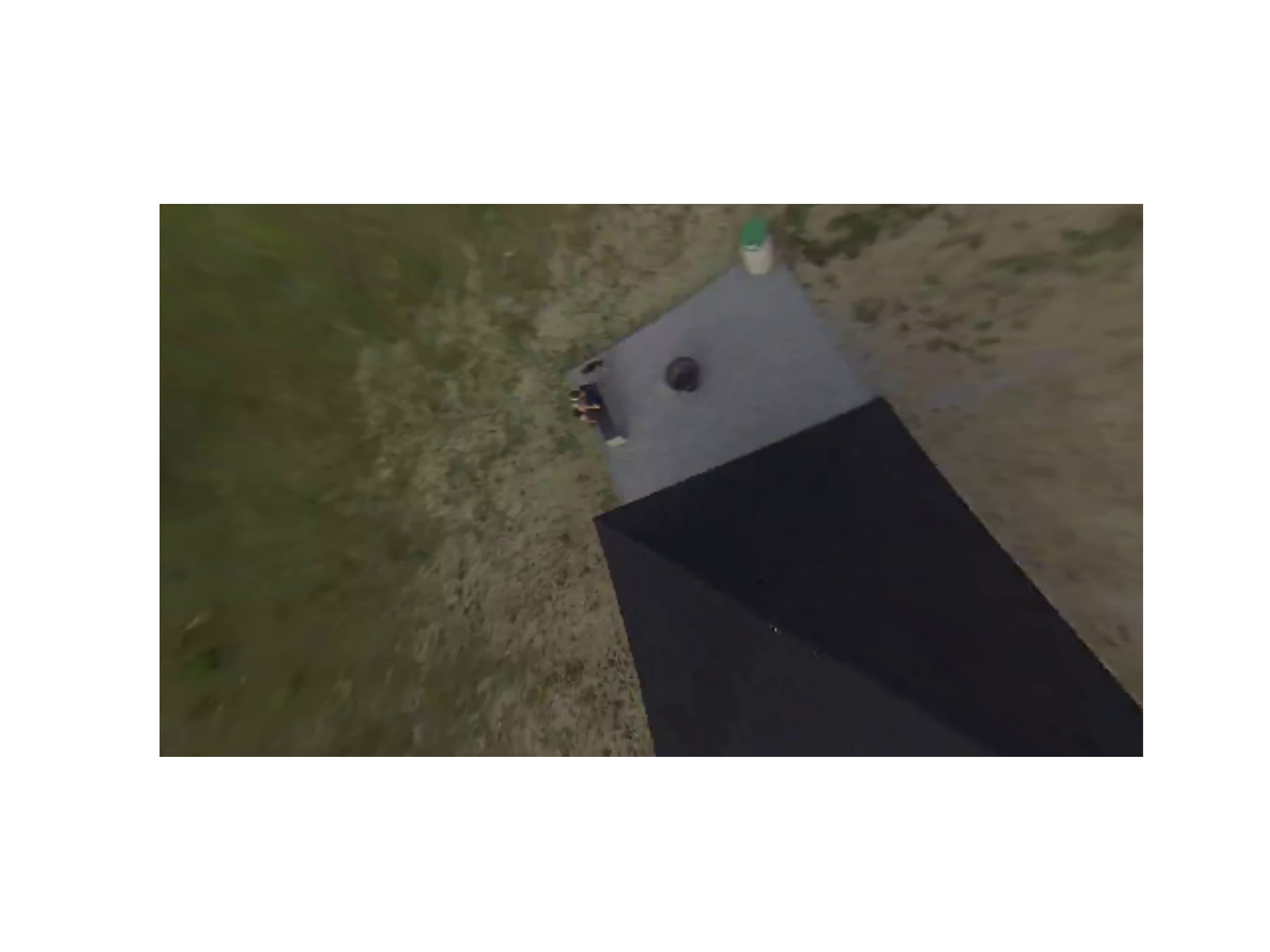}
    \includegraphics[trim={2cm 3cm 2cm 2cm},clip,width=0.30\columnwidth]{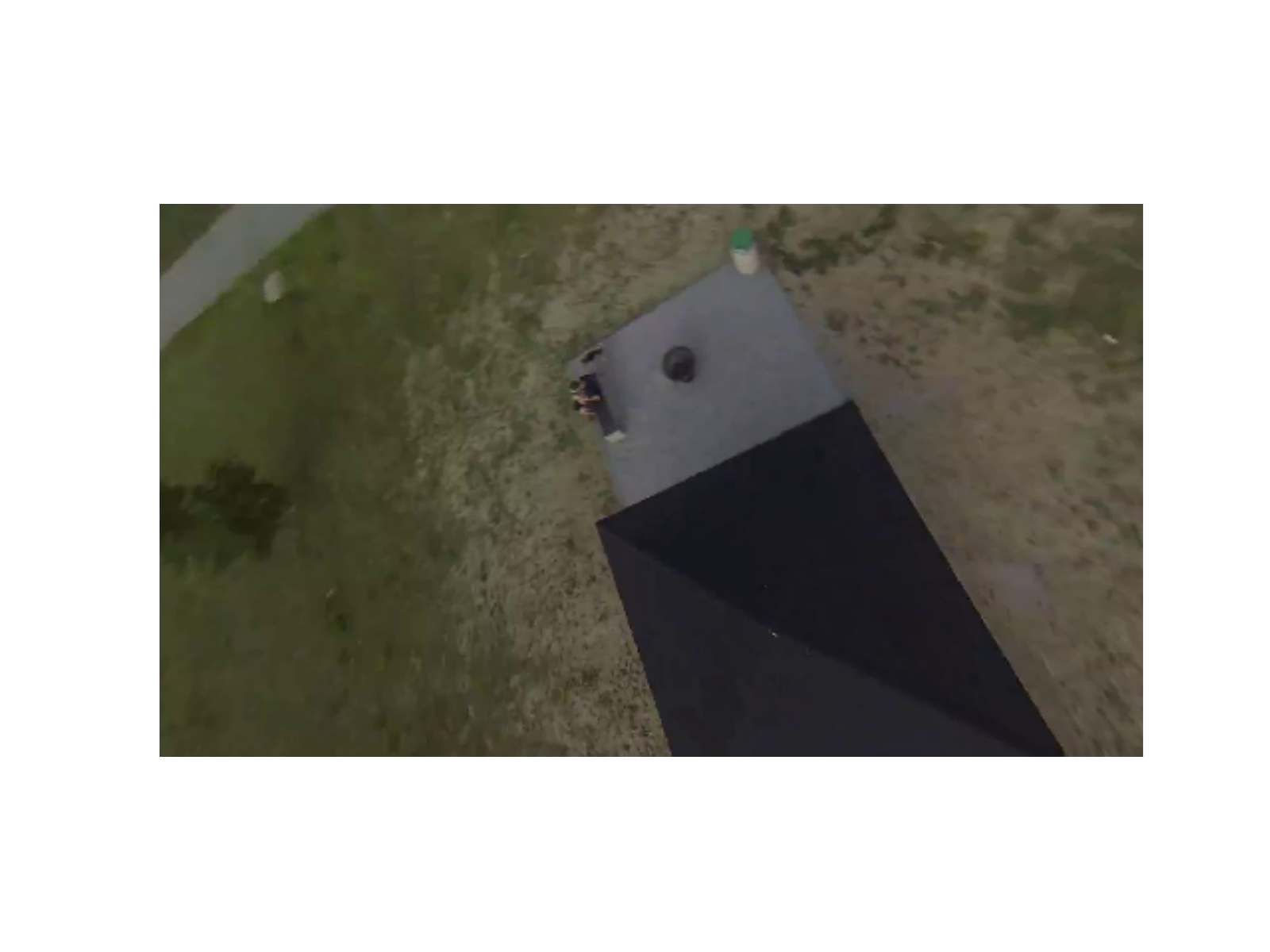}\\
    \caption{Right image shows tentative matches (black), inliers found using P4Pfr followed by R7Pfr (cyan) or R6P (blue) and the inliers after subsequent local optimization (LO) of the R7Pfr result (green) and the R6P result (red). The middle image and right image shows the RS and distortion removal using the R6P parameters after LO and the R7Pfr parameters after LO respectively.}
    \label{fig:real_inliers_undist}
    \vspace*{-0.5\baselineskip}
\end{figure}

The mean number of inliers on real data is summarized in Table~\ref{tab:real_inliers} and qualitative evaluation of image undistortion is shown in Figure~\ref{fig:real_inliers_undist}. Camera center precision is evaluated quantitatively in Table~\ref{tab:camera_centers} and qualitatively in supplementary material. Our solvers achieve overall better performance in terms of number of RANSAC inliers and R7Pfr followed by local optimization provides the best results in all cases. The estimated radial distortion and camera motion is significantly better than that of the baseline methods and can be readily used to remove both radial and RS distortion as shown in Figures~\ref{fig:real_inliers_undist} and~\ref{fig:nimrod_undist}. Relative focal length error compared to ground truth available in dataset Gopro drone 1 in Table~\ref{tab:focals} shows a significant improvement when using R7Pfr solver.
More  synthetic and real experiments are included in the supplementary material.
\begin{table}[t]
    \centering
        \caption{Mean relative errors of estimated focal length w.r.t the ground truth focal length $f_{gt} = 800px$ for the Gopro  datasets for our method and the baseline and the locally optimized variants (LO).}  
    \begin{tabular}{c|c|c|c|c}
    Dataset & P4Pfr+R6P & P4Pfr+R6P+LO & {\bf P4Pfr+R7Pfr} & {\bf P4Pfr+R7Pfr+LO} \\ \hline
    Gopro drone 1 & $10\%$ & $6.5\%$ & {\bf 2.37}\% & {\bf 1.5}\% \\ \hline
    Gopro rollerc. & $2.25\%$ & $1.75\%$ &  {\bf 1.75}\% & {\bf 1.5}\%
    \end{tabular}
    \label{tab:focals}
    \vspace*{-0.5\baselineskip}
\end{table}

\section{Conclusion}
We address the problem of absolute pose estimation of an uncalibrated RS camera, and present the first minimal solutions for the problem. Our two new minimal solvers are developed under the same computational scheme by combining an iterative scheme originally designed for calibrated RS cameras with fast generalized eigenvalue and efficient \gb solvers for specific polynomial equation systems. The R7Pf solver estimates the absolute pose of a RS camera with unknown focal length from 7 point correspondences. The R7Pfr solver estimates the absolute pose of a RS camera with unknown focal length and unknown radial distortion; also from 7 point correspondences. Our experiments demonstrate the accuracy of our new solvers.

\renewcommand{\thesection}{\Alph{section}}
\setcounter{section}{0}

\section{Appendix}
\subsection{Additional synthetic experiments}
In this section we show additional evaluation of the proposed algorithms on synthetic data.
We show two more experiments demonstrating the practical advantages of using R7Pf and R7Pfr. 

In these experiments we gradually increased rotational and translational velocity to the same values as in the experiments in Figure 1 in the main paper, e.g. rotation velocity up to 30 degrees/frame and relative translational velocity up to 1/10 of the camera distance from the scene per the duration of a frame. The difference from the experiments in the main paper is that this time the camera orientation is not set to identity and, therefore, R7Pf and R7Pfr have to be initialized by an initial rotation. We use the output rotation from P4Pf and P4Pfr to initialize R7Pf and R7Pfr respectively. 

\begin{figure}
    \centering
    \includegraphics[width=0.32\columnwidth]{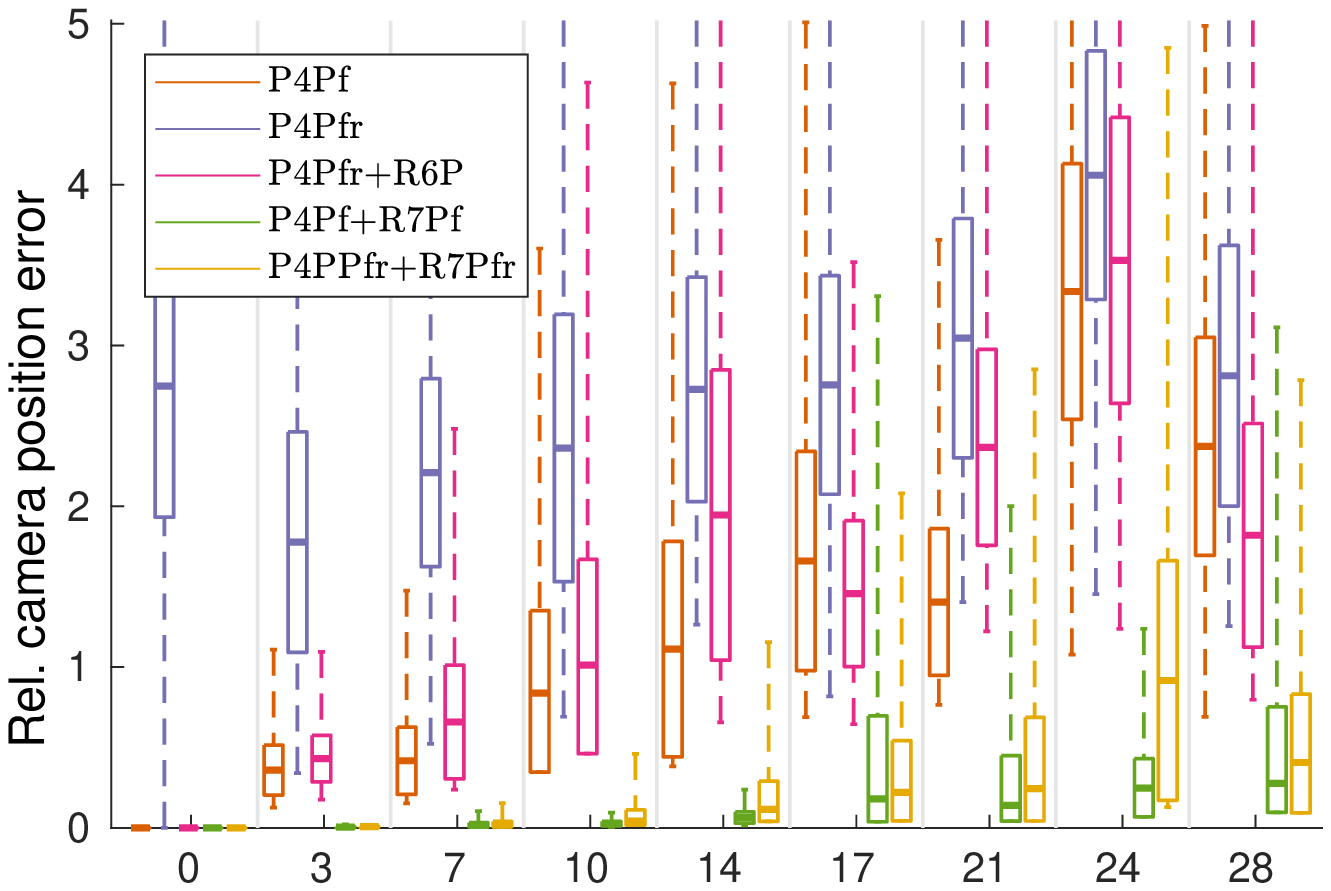}
    \includegraphics[width=0.32\columnwidth]{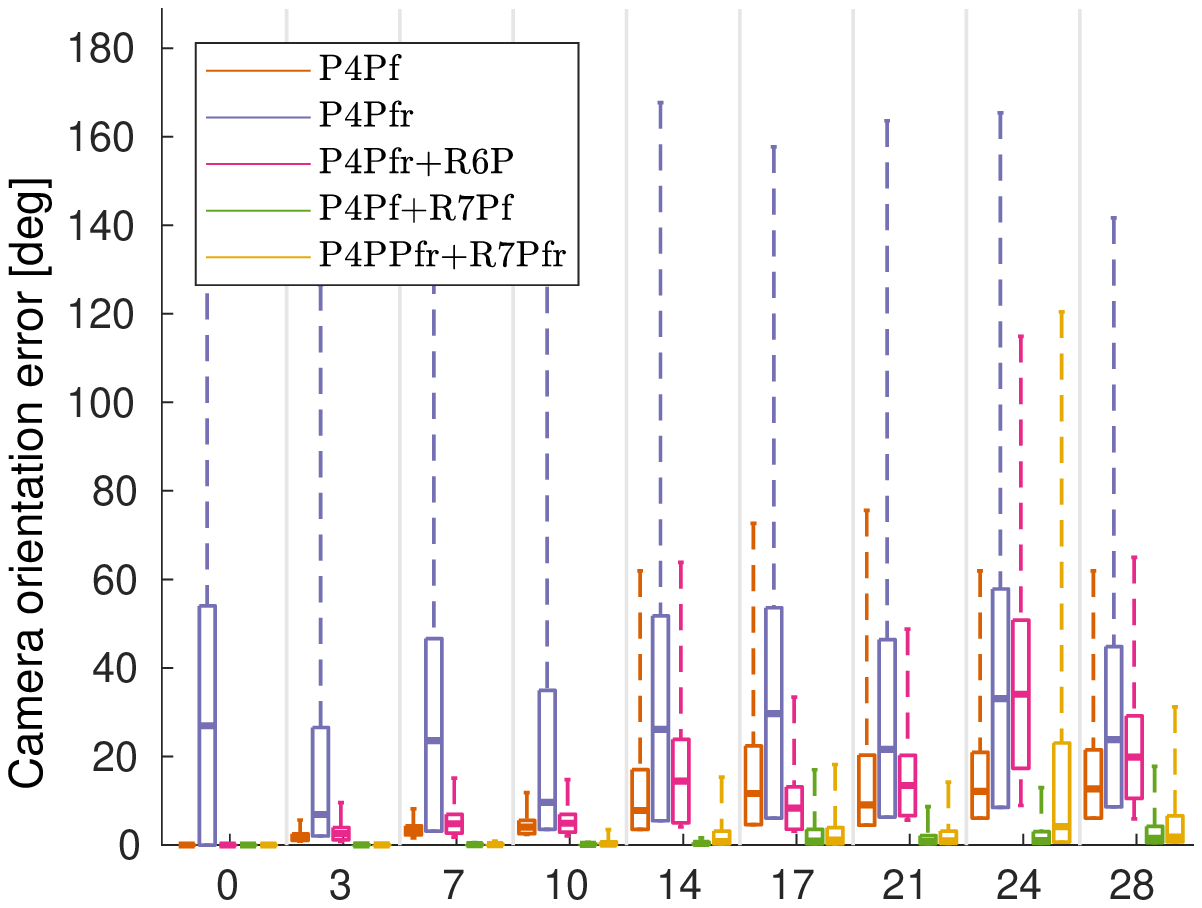}
    \includegraphics[width=0.32\columnwidth]{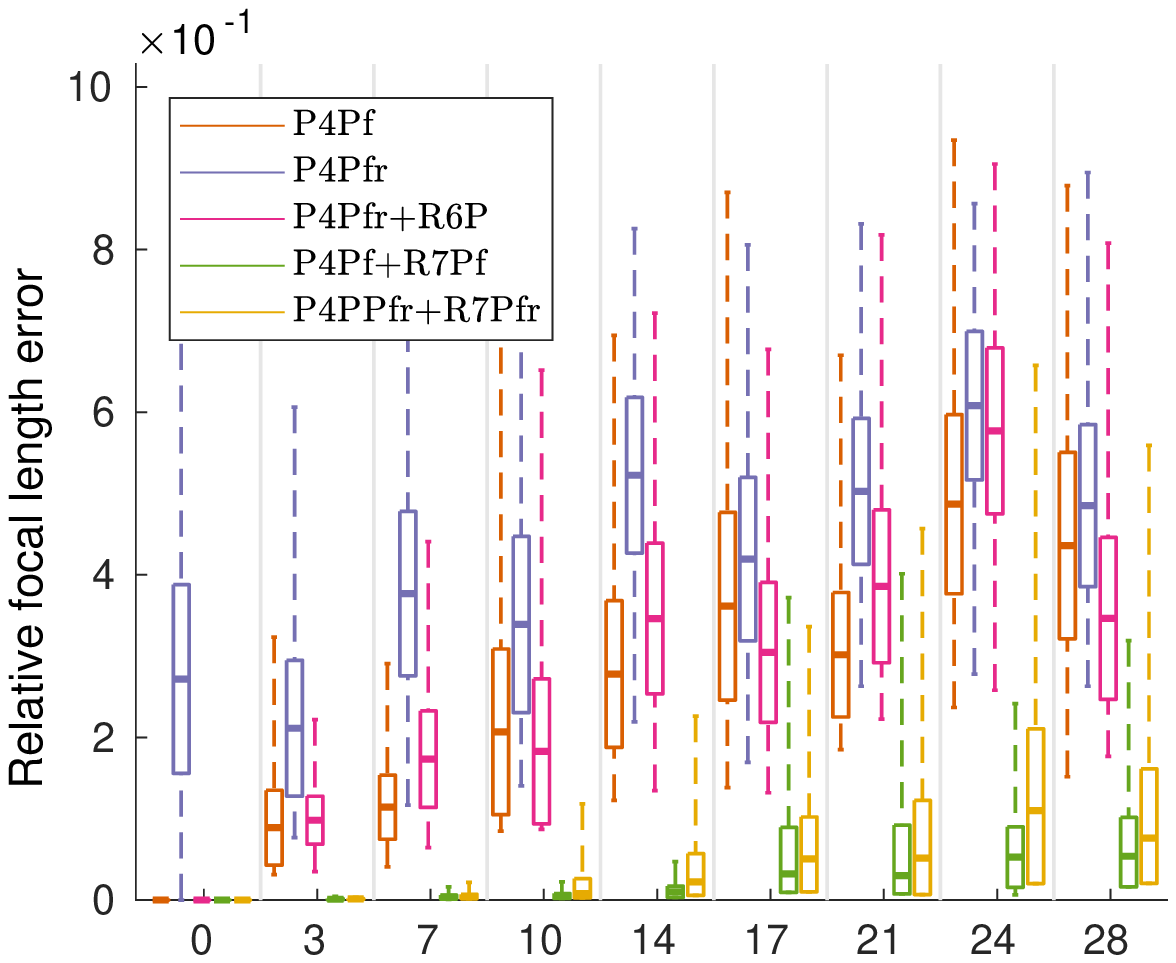}
    \caption{Performance of R7Pf and R7Pfr on data with increasing RS distortion and unknown focal length, when the initial orientation is initialized by P4Pf and P4Pfr respectively.}
    \label{fig:inc_wt_noradial}
\end{figure}

\begin{figure}
    \centering
    \includegraphics[width=0.32\columnwidth]{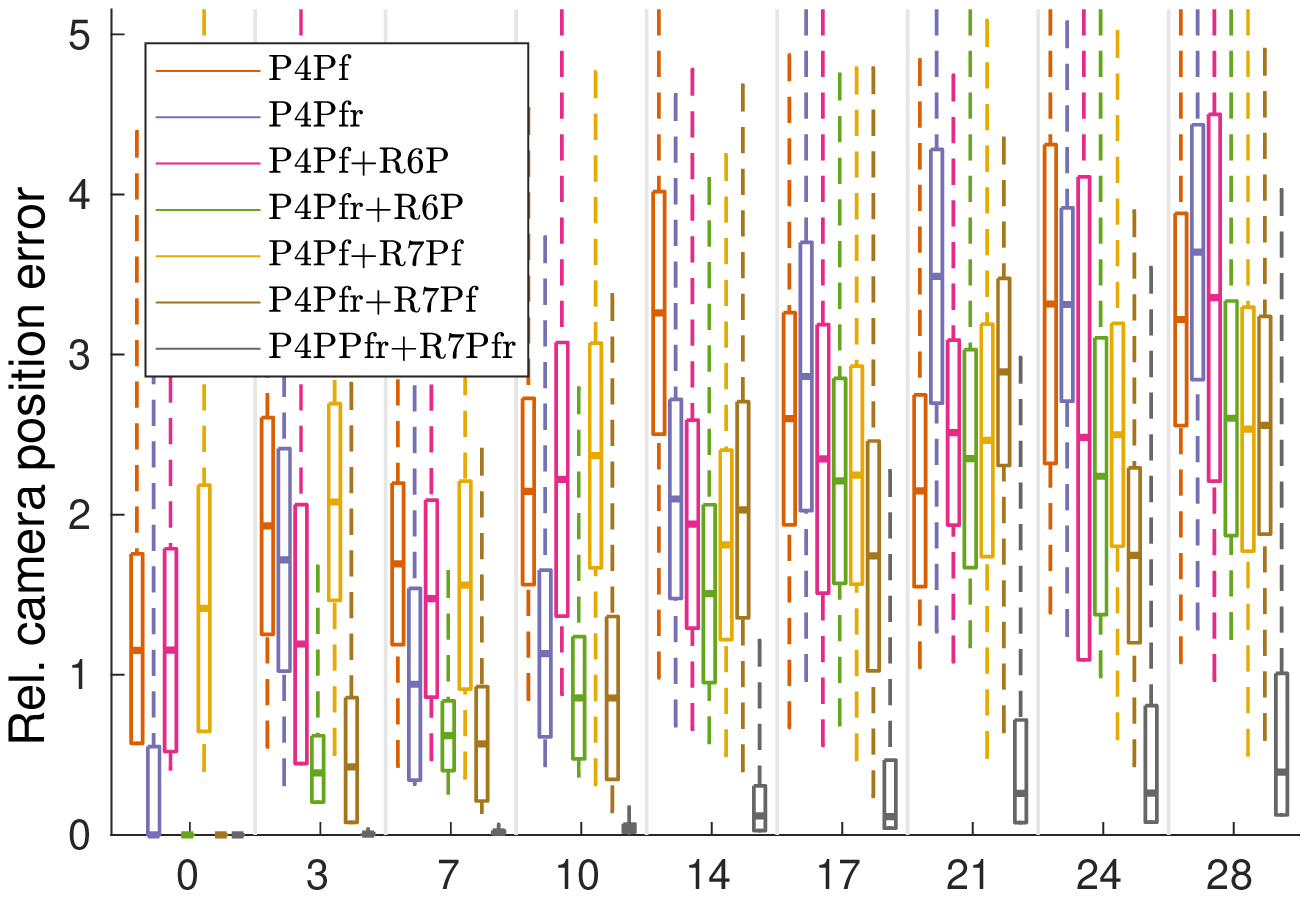}
    \includegraphics[width=0.32\columnwidth]{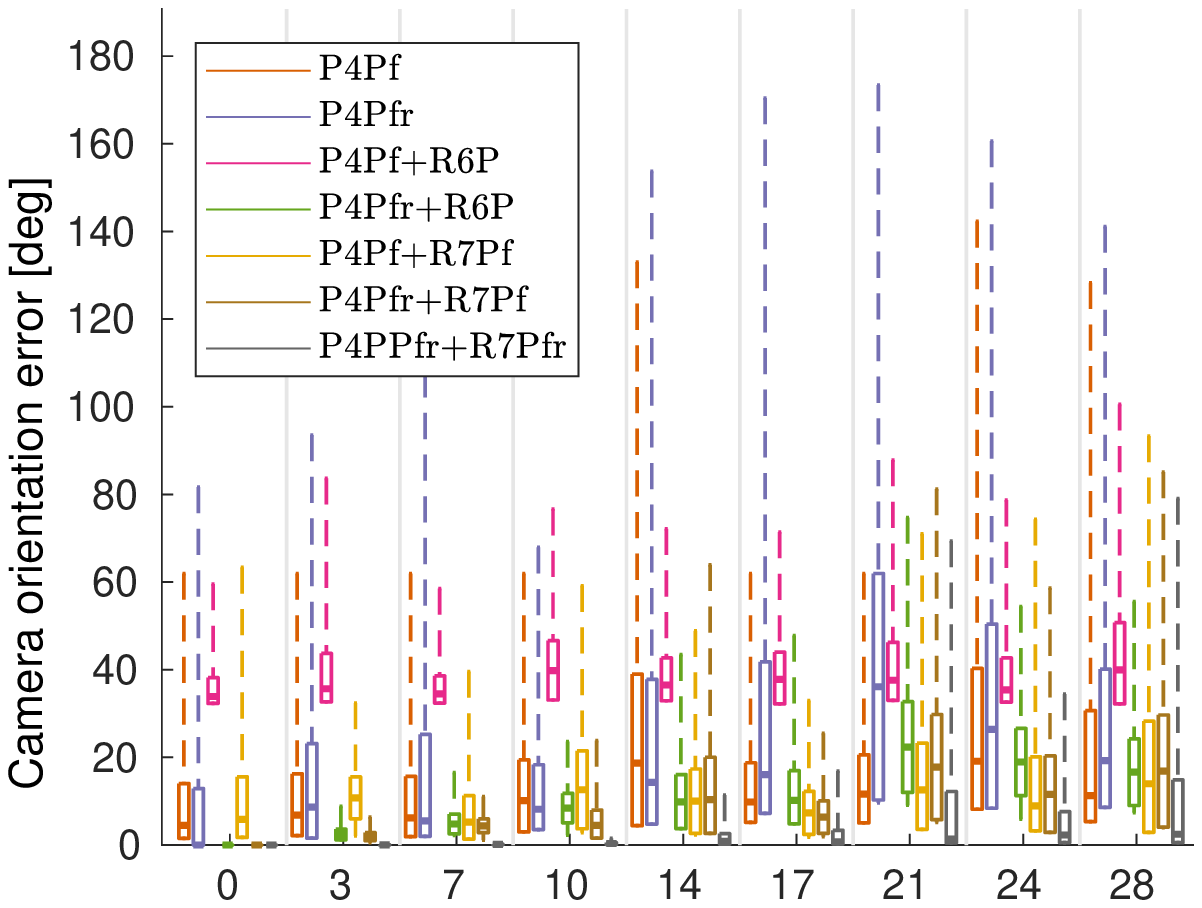}
    \includegraphics[width=0.32\columnwidth]{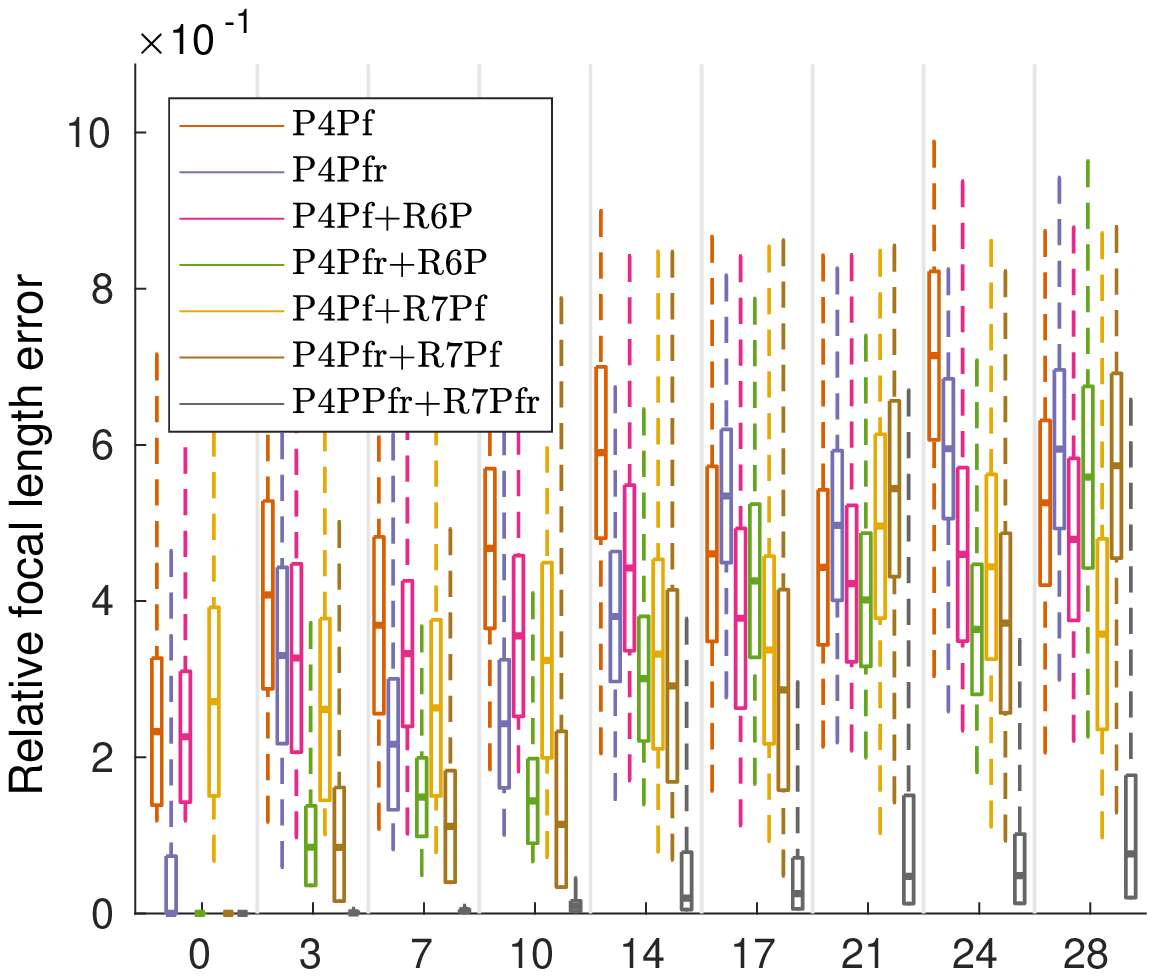}
    \caption{Performance of R7Pf and R7Pfr on data with increasing RS distortion, moderate radial distortion and unknown focal length, when the initial orientation is initialized by P4Pf and P4Pfr respectively.}
    \label{fig:inc_wt_radial}
\end{figure}

The data in the first experiment was generated without radial distortion whereas in the second experiment we used a fixed radial distortion of about half the maximum value of the one used in experiment in Figure 3 in the main paper.
Figure~\ref{fig:inc_wt_noradial} shows how R7P and R7Pfr should behave in a practical scenario with a moderate RS distortion, unknown focal length and no radial distortion and Figure~\ref{fig:inc_wt_radial} shows the case for radial distortion.

We can see that without radial distortion, the initialization by both P4Pf and P4Pfr is good enough to ensure R7Pf and R7Pfr provide a significantly better camera pose and focal length than the existing solutions. The P4Pfr+R7Pfr is significantly less stable on data without radial distortion, which indicates that the RS effect is being explained partially by the radial distortion. As expected on data with radial distoriton, P4Pf+R7Pf performs significantly poorer which indicates that radial distortion present in the image is being explained by some RS distortions, similar effect as with R7Pfr on non-distorted data. This is also visible in the extremely poor result of P4Pf+R6P. R7Pf initialized by P4Pfr on average outperforms P4Pfr+R6P, but it is clear from the results of both, that the radial distortion estimated by a solver without RS model (P4Pfr) is poor. R7Pfr provides the best performance and significantly outperforms all alternatives.

\subsection{Qualitative trajectory evaluation}
\begin{figure}
    \centering
    \includegraphics[angle=90,trim={22cm 7cm 20cm 7cm},clip,width=\columnwidth]{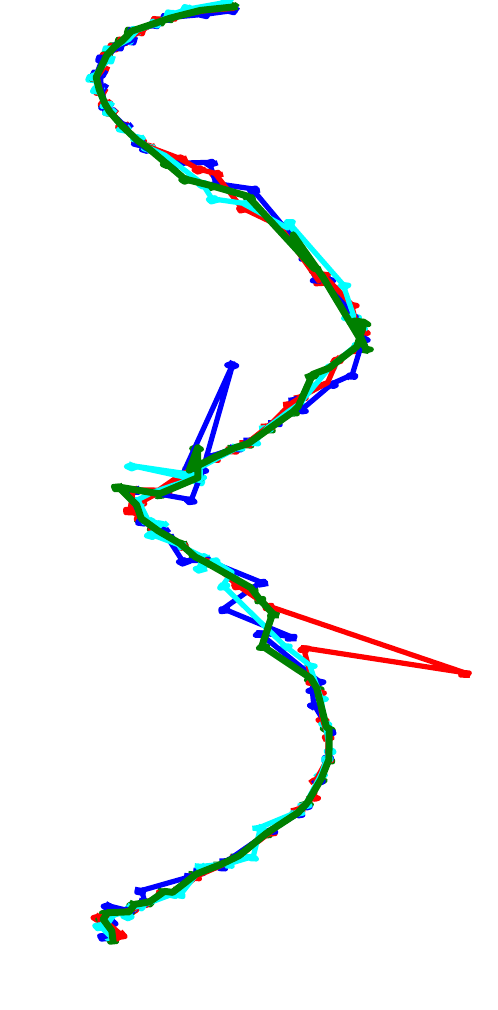}
    \includegraphics[angle=90,trim={8cm 5cm 7cm 5cm},clip,width=0.3\columnwidth]{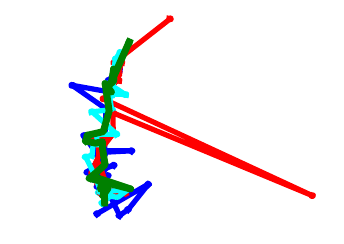}
    \caption{Reconstructed camera trajectories of the Gopro rollercoaster dataset (top) and Gopro drone 2 (bottom). The P4Pfr+R7Pfr (cyan) and P4Pfr+R7Pfr+LO (green) provide much more stable camera path than P4Pfr+R6P (blue) and P4Pfr+R6P+LO (red) in the critical places where camera motion is high.}
    \label{fig:traj}
\end{figure}

Here we show an example of the camera poses obtained by the compared algorithms. Figure~\ref{fig:traj} shows the camera centers calculated by P4Pfr+R7Pfr (cyan), \\ P4Pfr+R7Pfr+LO (green), P4Pfr+R6P (blue) and P4Pfr+R6P+LO (red) connected by lines which form a continuous trajectory of a drone performing a fast maneuver (bottom) and a rollercoaster performing a helix motion (top). One can observe that our solutions provide significantly more stable pose especially during fast motions. The baseline algorithms are prone to providing completely wrong pose at multiple occasions and overall suffer from lower accuracy caused by the lower number of detected inliers as well as interplay of the RS and radial distortion parameters. 

\section*{Acknowledgement}
T. Pajdla was supported by the European Regional Development Fund under IMPACT reg.\ no.\ CZ.02.1.01/0.0/0.0/15 003/0000468), EU H2020 No.~856994 ARtwin and EU H2020 No.~ 871245 SPRING Projects. ZK was supported by OP RDE project International Mobility of Researchers MSCA-IF at CTU Reg. No. CZ.02.2.69/0.0/0.0/$17\_050$/0008025 and OP VVV project Research Center for Informatics Reg. No. CZ.02.1.01/0.0/0.0/$16\_019$/0000765.

\clearpage
%
%
\bibliographystyle{splncs04}
\bibliography{egbib}
\end{document}